\documentclass[10pt,journal,compsoc]{IEEEtran}

\ifCLASSOPTIONcompsoc
  \usepackage[nocompress]{cite}
\else
  \usepackage{cite}
\fi

\usepackage{times, epsfig, graphicx, amsmath, amssymb, color}
\usepackage{hyperref}
\usepackage{booktabs, subcaption, multirow, tikz}
\usepackage{definitions}
\usepackage[normalem]{ulem}
\newcommand\del{\bgroup\markoverwith{\textcolor{red}{\rule[0.5ex]{2pt}{1.2pt}}}\ULon}

\begin{document}
%
% paper title
% \title{\methodname\\Real-time Instance Segmentation}
%\title{\chong{\methodname++\\Better and Still Real-time\\Instance Segmentation}}
\title{\methodname++\\Better Real-time Instance Segmentation}
%
%
% author names and IEEE memberships
\author{Daniel~Bolya\IEEEauthorrefmark{1},
        Chong~Zhou\IEEEauthorrefmark{1},
        Fanyi~Xiao,
        and~Yong~Jae~Lee
%\thanks{\IEEEauthorrefmark{1} The first two authors contributed equally to this work.}
\IEEEcompsocitemizethanks{
\IEEEcompsocthanksitem * The first two authors contributed equally to this work.
\IEEEcompsocthanksitem This work was performed at the Computer Vision Lab, Department of Computer Science, University of California, Davis, CA 95616 USA.\protect\\
E-mail: \{dbolya, cczhou, fyxiao, yongjaelee\}@ucdavis.edu.}
}

% The paper headers
\markboth{Journal of \LaTeX\ Class Files,~Vol.~X, No.~Y, December~2019}%
{Shell \MakeLowercase{\textit{et al.}}: Bare Demo of IEEEtran.cls for Computer Society Journals}

% The only time the second header will appear is for the odd numbered pages
% after the title page when using the twoside option.
% 
% *** Note that you probably will NOT want to include the author's ***
% *** name in the headers of peer review papers.                   ***
% You can use \ifCLASSOPTIONpeerreview for conditional compilation here if
% you desire.

% As a general rule, do not put math, special symbols or citations
% in the abstract or keywords.

\IEEEtitleabstractindextext{%
\begin{abstract}
We present a simple, fully-convolutional model for real-time ($>30$ fps) instance segmentation that achieves competitive results on MS COCO evaluated on a single Titan Xp, which is significantly faster than any previous state-of-the-art approach. Moreover, we obtain this result after training on \textbf{only one GPU}. We accomplish this by breaking instance segmentation into two parallel subtasks: (1) generating a set of prototype masks and (2) predicting per-instance mask coefficients. Then we produce instance masks by linearly combining the prototypes with the mask coefficients. We find that because this process doesn't depend on repooling, this approach produces very high-quality masks and exhibits temporal stability for free. Furthermore, we analyze the emergent behavior of our prototypes and show they learn to localize instances on their own in a translation variant manner, despite being fully-convolutional. We also propose Fast NMS, a drop-in 12 ms faster replacement for standard NMS that only has a marginal performance penalty. Finally, by incorporating deformable convolutions into the backbone network, optimizing the prediction head with better anchor scales and aspect ratios, and adding a novel fast mask re-scoring branch, our \methodname++ model can achieve $34.1$ mAP on MS COCO at $33.5$ fps, which is fairly close to the state-of-the-art approaches while still running at real-time.
%\chong{Finally, with the upgrade in the backbone detector, by incorporating deformable convolution and optimizing the prediction head, and a novel fast mask re-scoring branch, our \methodname++ model can achieve $34.1$ mAP on MS COCO at $33.5$ fps, which is fairly close to the state-of-the-art approaches while still running at real-time.}
\end{abstract}

% Note that keywords are not normally used for peerreview papers.
}
% https://ieeecs-media.computer.org/assets/pdf/taxonomy.pdf

% make the title area
\maketitle

\IEEEdisplaynontitleabstractindextext
% \IEEEdisplaynontitleabstractindextext has no effect when using
% compsoc or transmag under a non-conference mode.

% For peer review papers, you can put extra information on the cover
% page as needed:
% \ifCLASSOPTIONpeerreview
% \begin{center} \bfseries EDICS Category: 3-BBND \end{center}
% \fi
%
% For peerreview papers, this IEEEtran command inserts a page break and
% creates the second title. It will be ignored for other modes.
\IEEEpeerreviewmaketitle

%%%%%%%%%%%% INTRODUCTION %%%%%%%%%%%%
\IEEEraisesectionheading{\section{Introduction}\label{sec:introduction}}
% \vspace*{-0.05in}

\emph{``Boxes are stupid anyway though, I'm probably a true believer in masks except I can't get YOLO to learn them.''}
% \vspace*{-0.05in}
\begin{flushright}
      -- Joseph Redmon, YOLOv3~\cite{yolov3}
\end{flushright}

\IEEEPARstart{W}{hat} would it take to create a real-time instance segmentation algorithm? Over the past few years, the vision community has made great strides in instance segmentation, in part by drawing on powerful parallels from the well-established domain of object detection. State-of-the-art approaches to instance segmentation like Mask R-CNN \cite{maskrcnn} and FCIS \cite{fcis} directly build off of advances in object detection like Faster R-CNN \cite{fasterrcnn} and R-FCN \cite{rfcn}. Yet, these methods focus primarily on performance over speed, leaving the scene devoid of instance segmentation parallels to real-time object detectors like SSD \cite{ssd} and YOLO~\cite{yolov2, yolov3}. In this work, our goal is to fill that gap with a fast, one-stage instance segmentation model in the same way that SSD and YOLO fill that gap for object detection.

    \begin{figure}[t!]
    \centering
    % \hspace*{-0.5cm} \includegraphics[trim=5 0 0 0, clip, width=0.43\textwidth]{figures/graphics/tradeoff.pdf}
    % \hspace*{-0.5cm} \includegraphics[trim=10 0 65 40, clip, width=0.5\textwidth]{figures/graphics/tradeoff.png}
    \hspace*{-0.5cm} \includegraphics[trim=10 0 60 40, clip, width=0.45\textwidth]{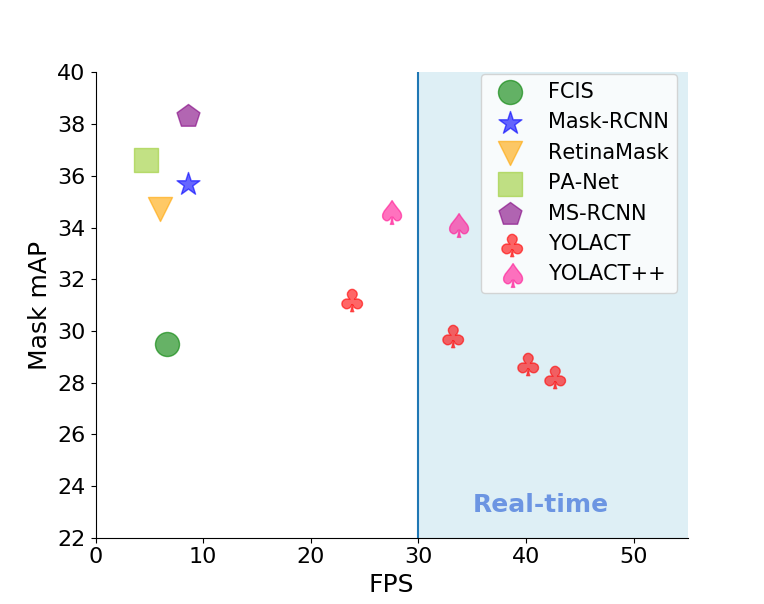}
    % \vspace{-0.12in}
    \caption{Speed-performance trade-off for various instance segmentation methods on COCO. To our knowledge, ours is the first \emph{real-time} (above 30 FPS) approach with over 30 mask mAP on COCO {\tt test-dev}. }
    % \vspace{-0.13in}
    \label{fig:speed_performance}
\end{figure}

However, instance segmentation is hard---much harder than object detection. One-stage object detectors like SSD and YOLO are able to speed up existing two-stage detectors like Faster R-CNN by simply removing the second stage and making up for the lost performance in other ways (e.g., strong data augmentation, anchor clustering, etc.). The same approach is not easily extendable, however, to instance segmentation. State-of-the-art two-stage instance segmentation methods depend heavily on \emph{feature localization} to produce masks. That is, these methods ``re-pool'' features in some bounding box region (e.g., via RoI-pool/align), and then feed these now localized features to their mask predictor. This approach is inherently sequential and is therefore difficult to accelerate. One-stage methods that perform these steps in parallel like FCIS do exist (e.g., \cite{fcis, xie-polarmask2019,benbarka-fouriernet2020}), but they require significant amounts of post-processing after localization, and thus are still far from real-time.

To address these issues, we propose \methodname{}\footnote{\textbf{Y}ou \textbf{O}nly \textbf{L}ook \textbf{A}t \textbf{C}oefficien\textbf{T}s}, a real-time instance segmentation framework that forgoes an explicit localization step. Instead, \methodname{} breaks up instance segmentation into two parallel tasks: (1) generating a dictionary of non-local \textit{prototype masks over the entire image}, and (2) predicting a set of \textit{linear combination coefficients per instance}. Then producing a full-image instance segmentation from these two components is simple: for each instance, linearly combine the prototypes using the corresponding predicted coefficients and then crop with a predicted bounding box. We show that by segmenting in this manner, \textit{the network learns how to localize instance masks on its own}, where visually, spatially, and semantically similar instances appear different in the prototypes.

Moreover, since the number of prototype masks is independent of the number of categories (e.g., there can be more categories than prototypes), \methodname{} learns a distributed representation in which each instance is segmented with a combination of prototypes that are shared across categories.  This distributed representation leads to interesting emergent behavior in the prototype space: some prototypes spatially partition the image, some localize instances, some detect instance contours, some encode position-sensitive directional maps (similar to those obtained by hard-coding a position-sensitive module in FCIS~\cite{fcis}), and most do a combination of these tasks (see Figure~\ref{fig:behavior}).

This approach also has several practical advantages. First and foremost, it's fast: because of its parallel structure and extremely lightweight assembly process, \methodname{} adds only a marginal amount of computational overhead to a one-stage backbone detector, making it easy to reach 30 fps even when using ResNet-101 \cite{resnet}; in fact, \emph{the entire mask branch takes only $\sim$5 ms to evaluate}. Second, masks are high-quality: since the masks use the full extent of the image space without any loss of quality from repooling, our masks for large objects are significantly higher quality than those of other methods (see Figure~\ref{fig:mask_quality}). Finally, it's general: the idea of generating prototypes and mask coefficients could be added to almost any modern object detector.

Our main contribution is the first real-time ($>30$ fps) instance segmentation algorithm with competitive results on the challenging MS COCO dataset~\cite{coco} (see Figure~\ref{fig:speed_performance}). In addition, we analyze the emergent behavior of \methodname{}'s prototypes and provide experiments to study the speed vs.~performance trade-offs obtained with different backbone architectures, numbers of prototypes, and image resolutions. We also provide a novel Fast NMS approach that is 12ms faster than traditional NMS with a negligible performance penalty. To further improve the performance of our model over our conference paper version \cite{yolact}, in Section~\ref{sec:yolact++}, we propose \methodname{}++. Specifically, we incorporate deformable convolutions~\cite{deformv1, deformv2} into the backbone network, which provide more flexible feature sampling and strengthening its capability of handling instances with different scales, aspect ratios, and rotations. Furthermore, we optimize the prediction heads with better anchor scale and aspect ratio choices for larger object recall. Finally, we also introduce a novel fast mask re-scoring branch, which results in a decent performance boost with only marginal speed overhead. These improvements are validated in Tables~\ref{tab:pred_head_ablation}, \ref{tab:yolact++_ablations}, and \ref{tab:dcn_ablations}.  Apart from these algorithm improvements over our conference paper \cite{yolact}, we also provide more qualitative results (Figure \ref{fig:more_qualitative}), a timing breakdown of each stage (Table \ref{tab:timing}), and real-time bounding box detection results (Table \ref{tab:detection}). %to demonstrate \methodname{}'s effectiveness in predicting adjacent instances in the same category and its performance as a real-time object detector, respectively.
%\chong{boosts the backbone detector by having a better and more flexible feature sampling policy and strengthening its capability of handling instances with different scales, aspect ratios, and rotations. Furthermore, we optimize the prediction heads with better anchor choice for larger box recall}

%\%chong{To further improve the performance of our conference paper \cite{yolact}, deformable convolution~\cite{deformv1, deformv2} and optimized prediction head are used in our \methodname++ model. Besides, \methodname++ also comes with a novel fast mask re-scoring branch, which is repooling-free to accommodate to \methodname{} resulting in a decent performance boost with marginal speed overhead. Furthermore, we provide more qualitative results (see Figure \ref{fig:more_qualitative}) and box results (see Table \ref{tab:detection}) to demonstrate \methodname{}'s effectiveness of predicting adjacent instances in the same category and its performance as a real-time object detector, respectively.} 

The code for \methodname{} and \methodname++ are both available at \href{https://github.com/dbolya/yolact}{https://github.com/dbolya/yolact}.

%%%%%%%%%%%% RELATED WORK %%%%%%%%%%%%
\section{Related Work}
% \vspace*{-0.1in}

\paragraph{Instance Segmentation}
Given its importance, a lot of research effort has been made to push instance segmentation \emph{accuracy}. Mask-RCNN~\cite{maskrcnn} is a representative two-stage instance segmentation approach that first generates candidate region-of-interests (ROIs) and then classifies and segments those ROIs in the second stage. Follow-up works try to improve its accuracy by e.g., enriching the FPN features~\cite{liu-panet2018} or addressing the incompatibility between a mask's confidence score and its localization accuracy~\cite{huang-msrcnn2018}. These two-stage methods require re-pooling features for each ROI and processing them with subsequent computations, which make them unable to obtain real-time speeds (30 fps) even when decreasing image size (see Table \ref{tab:accelerated_baselines}).

One-stage instance segmentation methods generate position sensitive maps that are assembled into final masks with position-sensitive pooling~\cite{dai-eccv2016,fcis} or combine semantic segmentation logits and direction prediction logits~\cite{chen-masklab2018}. Though conceptually faster than two-stage methods, they still require repooling or other non-trivial computations (e.g.,~mask voting). This severely limits their speed, placing them far from real-time. In contrast, our assembly step is much more lightweight (only a linear combination) and can be implemented as one GPU-accelerated matrix-matrix multiplication, making our approach very fast.

Finally, some methods first perform semantic segmentation followed by boundary detection~\cite{kirillov-cvpr2017}, pixel clustering~\cite{bai-cvpr2017,liang-pami2018}, CRF inference~\cite{arnab-cvpr2017}, or learn an embedding to form instance masks~\cite{newell-nips2017,harley-iccv2017,de-arxiv2017,fathi-arxiv2017}. Again, these methods have multiple stages and/or involve expensive clustering procedures, which limits their viability for real-time applications.

    \begin{figure*}
    \centering
    \includegraphics[trim=0 0 0 0, clip, width=0.95\textwidth]{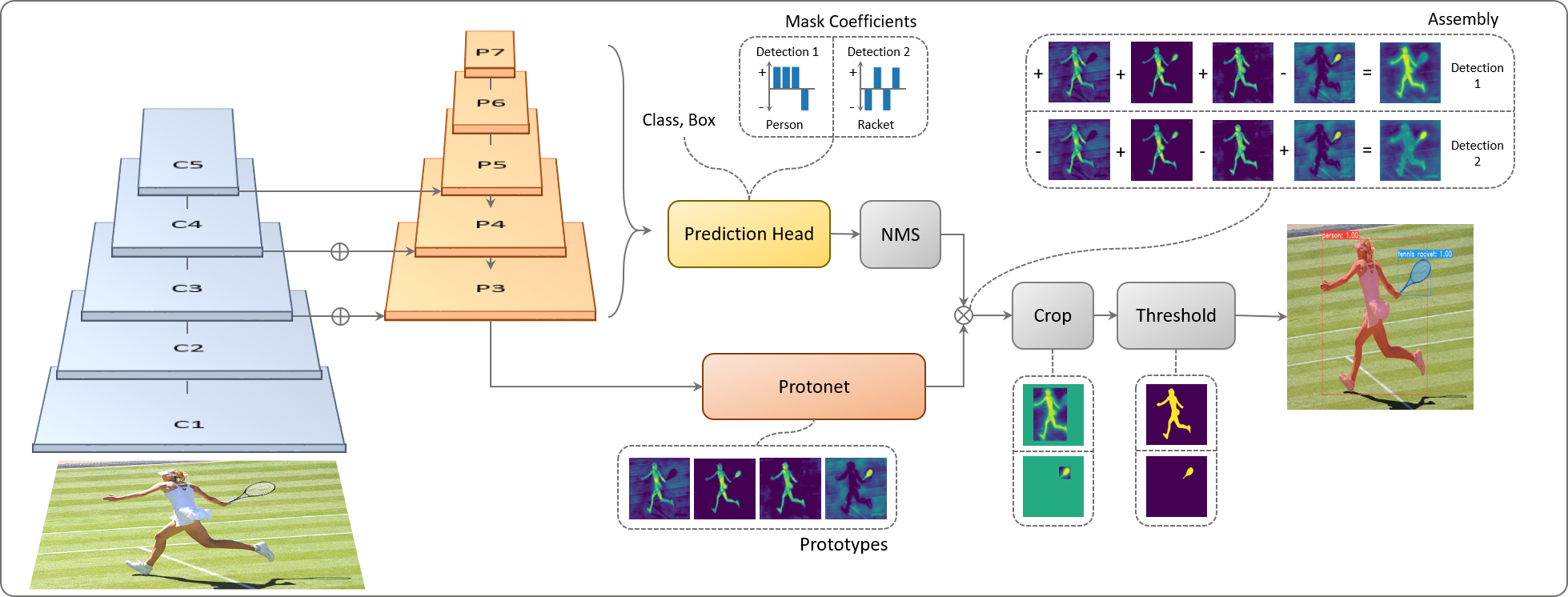}
    % \vspace{-0.14in}
    \caption{\inlinesection{\methodname{} Architecture} Blue/yellow indicates low/high values in the prototypes, gray nodes indicate functions that are not trained, and $k=4$ in this example. We base this architecture off of RetinaNet~\cite{retinanet} using ResNet-101 + FPN.}
    % \vspace{-0.11in}
    \label{fig:concept}
\end{figure*}

\paragraph{Real-time Instance Segmentation}
While real-time object detection~\cite{ssd,yolov1,yolov2,yolov3}, and semantic segmentation~\cite{segnet,treml2016sq,enet,blitznet,zhao2018icnet} methods exist, few works have focused on real-time instance segmentation.  Straight to Shapes~\cite{straight2shapes} and Box2Pix~\cite{box2pix} can perform instance segmentation in real-time (30 fps on Pascal SBD 2012~\cite{pascalvoc,sbd} for Straight to Shapes, and 10.9 fps on Cityscapes~\cite{cityscape} and 35 fps on KITTI~\cite{KITTI} for Box2Pix), but their accuracies are far from that of modern baselines. While \cite{neven2019instance} substantially improves instance segmentation accuracy over these prior methods, it runs only at 11 fps on Cityscapes. In fact, Mask R-CNN~\cite{maskrcnn} remains one of the fastest instance segmentation methods on semantically challenging datasets like COCO~\cite{coco} (13.5 fps on $550^2$ px images; see Table~\ref{tab:accelerated_baselines}).
%\chong{\cite{neven2019instance}, on the other hand, has a competitive performance and runs at 11 fps on Cityscapes, but it is still not even close to 30 fps.}

\paragraph{Prototypes}
Learning prototypes (aka vocabulary/codebook) has been extensively explored in computer vision. Classical representations include textons~\cite{textons} and visual words~\cite{sivic-iccv2003}, with advances made via sparsity and locality priors~\cite{yang-tip2010,wang-cvpr2010,zhang-iccv2013}. Others have designed prototypes for object detection~\cite{agarwal-eccv2002,yu-bmvc2007,ren-cvpr2013}. Though related, these works use prototypes to represent features, whereas we use them to assemble masks for instance segmentation. Moreover, we learn prototypes that are specific to each image, rather than global prototypes shared across the entire dataset like in~\cite{arnab-cvpr2017}. Also, unlike the ``shape priors'' defined in~\cite{arnab-cvpr2017}, which are fixed shape primitives, our prototypes are per-image feature maps that all masks can draw from.

%%%%%%%%%%%% METHOD %%%%%%%%%%%%
\section{\methodname}

Our goal is to add a mask branch to an existing one-stage object detection model in the same vein as Mask R-CNN \cite{maskrcnn} does to Faster R-CNN \cite{fasterrcnn}, but without an explicit feature localization step (e.g., feature repooling). To do this, we break up the complex task of instance segmentation into two simpler, parallel tasks that can be assembled to form the final masks. The first branch uses an FCN \cite{fcn} to produce a set of image-sized ``prototype masks'' that do not depend on any one instance. The second adds an extra head to the object detection branch to predict a vector of ``mask coefficients'' for each anchor that encode an instance's representation in the prototype space. Finally, for each instance that survives box-based NMS, we construct a mask for that instance by linearly combining the work of these two branches.

\paragraph{Rationale}
We perform instance segmentation in this way primarily because masks are spatially coherent; i.e., pixels close to each other are likely to be part of the same instance. While a convolutional (\textit{conv}) layer naturally takes advantage of this coherence, a fully-connected (\textit{fc}) layer does not. That poses a problem, since one-stage object detectors produce class and box coefficients for each anchor as an output of an \textit{fc} layer.\footnotemark ~Two stage approaches like Mask R-CNN get around this problem by using a localization step (e.g., RoI-Align), which preserves the spatial coherence of the features while also allowing the mask to be a \textit{conv} layer output. However, doing so requires a significant portion of the model to wait for a first-stage RPN to propose localization candidates, inducing a significant speed penalty.

\footnotetext{To show that this is an issue, we develop an ``\textit{fc}-mask'' model that produces masks for each anchor as the reshaped output of an \textit{fc} layer. As our experiments in Table \ref{tab:accelerated_baselines} show, simply adding masks to a one-stage model as \textit{fc} outputs only obtains 20.7 mAP and is thus very much insufficient.}

Thus, we break the problem into two parallel parts, making use of \textit{fc} layers, which are good at producing semantic vectors, and \textit{conv} layers, which are good at producing spatially coherent masks, to produce the ``mask coefficients'' and ``prototype masks'', respectively. Then, because prototypes and mask coefficients can be computed independently, the computational overhead over that of the backbone detector comes mostly from the assembly step, which can be implemented as a single matrix multiplication.  In this way, we can maintain spatial coherence in the feature space while still being one-stage and \emph{fast}.

\subsection{Prototype Generation}
\begin{figure}
    \centering

    \includegraphics[trim=135 230 230 190, clip, width=0.35\textwidth]{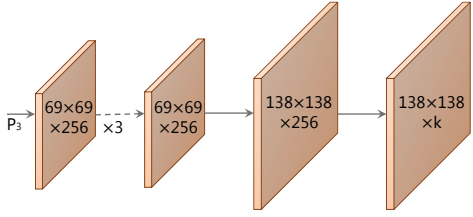}
    % \vspace{-0.06in}
    
    \caption{\inlinesection{Protonet Architecture} The labels denote feature size and channels for an image size of $550 \times 550$. Arrows indicate $3\times 3$ \textit{conv} layers, except for the final \textit{conv} which is $1 \times 1$. The increase in size is an upsample followed by a \textit{conv}. Inspired by the mask branch in \cite{maskrcnn}. }
    % \vspace{-0.1in}
    \label{fig:protonet}
\end{figure}

% \vspace*{-0.01in}
The prototype generation branch (protonet) predicts a set of $k$ prototype masks for the entire image. We implement protonet as an FCN whose last layer has $k$ channels (one for each prototype) and attach it to a backbone feature layer (see~Figure~\ref{fig:protonet} for an illustration). While this formulation is similar to standard semantic segmentation, it differs in that we exhibit no explicit loss on the prototypes. Instead, all supervision for these prototypes comes from the final mask loss after assembly.

We note two important design choices: taking protonet from deeper backbone features produces more robust masks, and higher resolution prototypes result in both higher quality masks and better performance on smaller objects. Thus, we use FPN \cite{fpn} because its largest feature layers ($P_3$ in our case; see Figure~\ref{fig:concept}) are the deepest. Then, we upsample it to one fourth the dimensions of the input image to increase performance on small objects.

Finally, we find it important for the protonet's output to be unbounded, as this allows the network to produce large, overpowering activations for prototypes it is very confident about (e.g., obvious background). Thus, we have the option of following protonet with either a {\tt ReLU} or no nonlinearity. We choose {\tt ReLU} for more interpretable prototypes.

\subsection{Mask Coefficients}
Typical anchor-based object detectors have two branches in their prediction heads: one branch to predict $c$ class confidences, and the other to predict $4$ bounding box regressors. For mask coefficient prediction, we simply add a third branch in parallel that predicts $k$ mask coefficients, one corresponding to each prototype. Thus, instead of producing $4 + c$ coefficients per anchor, we produce $4 + c + k$.

Then for nonlinearity, we find it important to be able to subtract out prototypes from the final mask. Thus, we apply {\tt tanh} to the $k$ mask coefficients, which produces more stable outputs over no nonlinearity. The relevance of this design choice is apparent in Figure \ref{fig:concept}, as neither mask would be constructable without allowing for subtraction.

\subsection{Mask Assembly}
% \vspace{-0.01in}
To produce instance masks, we combine the work of the prototype branch and mask coefficient branch, using a linear combination of the former with the latter as coefficients. We then follow this by a sigmoid nonlinearity to produce the final masks. These operations can be implemented efficiently using a single matrix multiplication and sigmoid:
    \begin{align} \vspace{-0.3in}
        M = \sigma{(P C^{T})} 
    \vspace{-0.2in} \end{align}
where $P$ is an $h\times w \times k$ matrix of prototype masks and $C$ is a $n \times k$ matrix of mask coefficients for $n$ instances surviving NMS and score thresholding. Other, more complicated combination steps are possible; however, we keep it simple (and fast) with a basic linear combination.

\paragraph{Losses} We use three losses to train our model: classification loss $L_{cls}$, box regression loss $L_{box}$ and mask loss $L_{mask}$ with the weights 1, 1.5, and 6.125 respectively. Both $L_{cls}$ and $L_{box}$ are defined in the same way as in~\cite{ssd}. Then to compute mask loss, we simply take the pixel-wise binary cross entropy between assembled masks $M$ and the ground truth masks $M_{gt}$: $L_{mask} = \text{BCE}(M, M_{gt})$.

\paragraph{Cropping Masks}
\label{sec:crop_mask}
We crop the final masks with the predicted bounding box during evaluation.  Specifically, we assign zero to pixels outside of the box region.  During training, we instead crop with the ground truth bounding box, and divide $L_{mask}$ by the ground truth box area to preserve small objects in the prototypes.

\subsection{Emergent Behavior}
Our approach might seem surprising, as the general consensus around instance segmentation is that because FCNs are equivariant with respect to input translations, the task needs position awareness added back in \cite{fcis}. Thus methods like FCIS \cite{fcis} and Mask R-CNN \cite{maskrcnn} try to explicitly add position awareness, whether it be by directional maps and position-sensitive repooling, or by putting the mask branch in the second stage so it does not have to deal with localizing instances. In our method, the only position awareness we add is to crop the final mask with the predicted bounding box. However, we find that our method also works without cropping for medium and large objects, so this is not a result of cropping. Instead, \methodname{} \textit{learns how to localize instances on its own} via different activations in its prototypes.

    \begin{figure}
    \centering
    
    \begin{tikzpicture}
        \draw (0, 0) node [inner sep=0] {\includegraphics[trim=20 85 40 75, clip, scale=0.32]{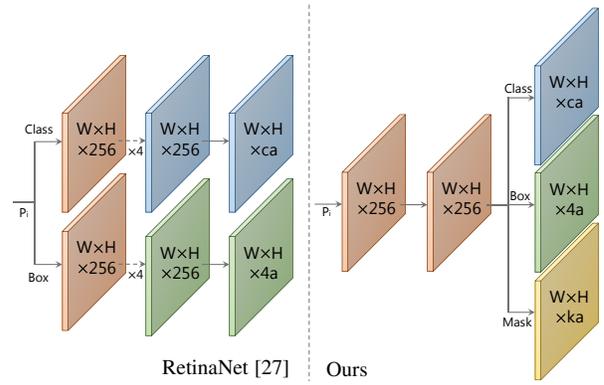}};
        \draw (-0.3, -2.4) node [anchor=east, font=\footnotesize] {RetinaNet \cite{retinanet}};
        \draw (-0.05, -2.4) node [anchor=west, font=\footnotesize] {Ours};
    \end{tikzpicture}
    %\vspace{-0.11in}
    \caption{\inlinesection{Head Architecture} We use a shallower prediction head than RetinaNet~\cite{retinanet} and add a mask coefficient branch. This is for $c$ classes, $a$ anchors for feature layer $P_i$, and $k$ prototypes. See Figure~\ref{fig:protonet} for a key.}
    %\vspace{-0.11in}
    \label{fig:pred_head}
\end{figure}
    
\begin{figure}[t!]
    \centering
    \includegraphics[width=0.46\textwidth]{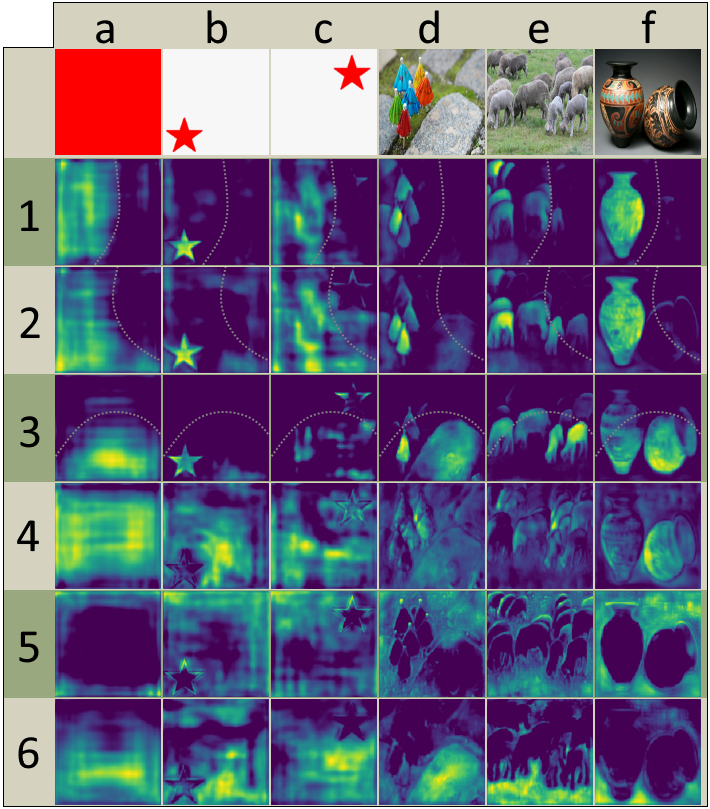}
    % \vspace{-0.1in}
    \caption{\inlinesection{Prototype Behavior} The activations of the same six prototypes (y axis) across different images (x axis). Prototypes 1-3 respond to objects to one side of a soft, implicit boundary (marked with a dotted line). Prototype 4 activates on the bottom-left of objects (for instance, the bottom left of the umbrellas in image d); prototype 5 activates on the background and on the edges between objects; and prototype 6 segments what the network perceives to be the ground in the image. These last 3 patterns are most clear in images d-f.}
    % \vspace{-0.11in}
    \label{fig:behavior}
    
    %The activations of the same six prototypes across different images. Prototypes 1, 4, and 5 are partition maps with boundaries clearly defined in image a, prototype 2 is a bottom-left directional map, prototype 3 segments out the background and provides instance contours, and prototype 6 segments out the ground.
\end{figure}

To see how this is possible, first note that the prototype activations for the solid red image (image a) in Figure \ref{fig:behavior} are actually not possible in an FCN without padding. Because a convolution outputs to a single pixel, if its input everywhere in the image is the same, the result everywhere in the \textit{conv} output will be the same. On the other hand, the consistent rim of padding in modern FCNs like ResNet gives the network the ability to tell how far away from the image's edge a pixel is. Conceptually, one way it could accomplish this is to have multiple layers in sequence spread the padded 0's out from the edge toward the center (e.g., with a kernel like $[1, 0]$). In practice, this leads to massive amount of effective padding for standard ConvNets like ResNets. For instance, ResNet 101 and ResNet 50 have 511px and 239px of padding in each direction respectively \cite{araujo2019computing}. In addition to the large receptive field (e.g., 1027 pixels for ResNet 101), this means ResNet, \textit{is inherently translation variant}, and our method makes heavy use of that property (images b and c exhibit clear translation variance).

We observe many prototypes to activate on certain ``partitions'' of the image. That is, they only activate on objects on one side of an implicitly learned boundary. In Figure \ref{fig:behavior}, prototypes 1-3 are such examples. By combining these partition maps, the network can distinguish between different (even overlapping) instances of the same semantic class; e.g., in image d, the green umbrella can be separated from the red one by subtracting prototype 3 from prototype 2.

Furthermore, being learned objects, prototypes are compressible. That is, if protonet combines the functionality of multiple prototypes into one, the mask coefficient branch can learn which situations call for which functionality. For instance, in Figure \ref{fig:behavior}, prototype 2 is a partitioning prototype but also fires most strongly on instances in the bottom-left corner. Prototype 3 is similar but for instances on the right. This explains why in practice, the model does not degrade in performance even with as low as $k=32$ prototypes (see Table~\ref{tab:num_proto}).

On the other hand, increasing $k$ is ineffective most likely because predicting coefficients is difficult. If the network makes a large error in even one coefficient, due to the nature of linear combinations, the produced mask can vanish or include leakage from other objects. Thus, the network has to play a balancing act to produce the right coefficients, and adding more prototypes makes this harder. In fact, we find that for higher values of $k$, the network simply adds redundant prototypes with small edge-level variations that slightly increase $AP_{95}$, but not much else.
%%%%%%%%%%%% BACKBONE DETECTOR %%%%%%%%%%%%
\section{Backbone Detector}

For our backbone detector we prioritize speed as well as feature richness, since predicting these prototypes and coefficients is a difficult task that requires good features to do well. Thus, the design of our backbone detector closely follows RetinaNet \cite{retinanet} with an emphasis on speed.

\paragraph{\methodname{} Detector}
We use ResNet-101 \cite{resnet} with FPN \cite{fpn} as our default feature backbone and a base image size of $550 \times 550$. We do not preserve aspect ratio in order to get consistent evaluation times per image, and at least on the almost-square COCO images, we don't observe any benefit for maintaining aspect ratio. Like RetinaNet, we modify FPN by not producing $P_2$ and producing $P_6$ and $P_7$ as successive $3 \times 3$ stride 2 \textit{conv} layers starting from $P_5$ (not $C_5$) and place 3 anchors with aspect ratios $[1, 1/2, 2]$ on each. The anchors of $P_3$ have areas of $24$ pixels squared, and every subsequent layer has double the scale of the previous (resulting in the scales $[24, 48, 96, 192, 384]$). For the prediction head attached to each $P_i$, we have one $3 \times 3$ \textit{conv} shared by all three branches, and then each branch gets its own $3 \times 3$ \text{conv} in parallel. Compared to RetinaNet, our prediction head design (see Figure \ref{fig:pred_head}) is more lightweight and much faster. We apply smooth-$L_1$ loss to train box regressors and encode box regression coordinates in the same way as SSD \cite{ssd}. To train class prediction, we use softmax cross entropy with $c$ positive labels and 1 background label, selecting training examples using OHEM \cite{ohem} with a 3:1 neg:pos ratio. Thus, unlike RetinaNet we do not use focal loss, which we found not to be viable in our situation. Finally, we do not do NMS during training, as one ground truth can match to multiple predictions. For each such positive prediction, we train both its box and mask.

With these design choices, we find that this backbone performs better and faster than SSD \cite{ssd} modified to use ResNet-101 \cite{resnet}, with the same image size.

%%%%%%%%%%%% OTHER IMPROVEMENTS %%%%%%%%%%%%
\section{Other Improvements}
We also discuss other improvements that either increase speed with little effect on performance or increase performance with no speed penalty.

\subsection{Fast NMS}
After producing bounding box regression coefficients and class confidences for each anchor, like most object detectors we perform NMS to suppress duplicate detections. In many previous works \cite{yolov2, yolov3, ssd, fasterrcnn, maskrcnn, retinanet}, NMS is performed sequentially. That is, for each of the $c$ classes in the dataset, sort the detected boxes descending by confidence, and then for each detection remove all those with lower confidence than it that have an IoU overlap greater than some threshold. While this sequential approach is fast enough at speeds of around 5 fps, it becomes a large barrier for obtaining 30 fps (for instance, a 10 ms improvement at 5 fps results in a 0.26 fps boost, while a 10 ms improvement at 30 fps results in a 12.9 fps boost).

To fix the sequential nature of traditional NMS, we introduce Fast NMS, a version of NMS where every instance can be decided to be kept or discarded in parallel. To do this, we simply allow already-removed detections to suppress other detections, which is not possible in traditional NMS. This relaxation allows us to implement Fast NMS entirely in standard GPU-accelerated matrix operations.
%This relaxation allows us to implement Fast NMS entirely in standard matrix operations available in most GPU-accelerated libraries.

%Then, we find which detections to remove by checking if there are any higher-scoring detections with a corresponding IoU greater than some threshold $t$.
To perform Fast NMS, we first compute a $c \times n\times n$ pairwise IoU matrix $X$ for the top $n$ detections sorted descending by score for each of $c$ classes. Batched sorting on the GPU is readily available and computing IoU can be easily vectorized. Then, we remove detections if there are any higher-scoring detections with a corresponding IoU greater than some threshold $t$. We efficiently implement this by first setting the lower triangle and diagonal of $X$ to 0: $X_{kij} = 0, ~\forall k, j, i \geq j,$
    % \begin{align} \label{eq:triu}
    %     X_{kij} = 0 \qquad \forall k, j, i \geq j
    % \end{align} 
which can be performed in one batched {\tt triu} call, and then taking the column-wise max:
    \begin{align} \label{eq:max}
        K_{kj} = \max_i(X_{kij}) \qquad \forall k, j
    \end{align} 
to compute a matrix $K$ of maximum IoU values for each detection. Finally, thresholding this matrix with $t$ ($K < t$) will indicate which detections to keep for each class.

Because of the relaxation, Fast NMS has the effect of removing slightly too many boxes. However, the performance hit caused by this is negligible compared to the stark increase in speed (see Table \ref{tab:nms}). In our code base, Fast NMS is 11.8 ms faster than a Cython implementation of traditional NMS while only reducing performance by 0.1 mAP. In the Mask R-CNN benchmark suite \cite{maskrcnn}, Fast NMS is 15.0 ms faster than their CUDA implementation of traditional NMS with a performance loss of only 0.3 mAP.

% \vspace{-0.01in}
\subsection{Semantic Segmentation Loss}
While Fast NMS trades a small amount of performance for speed, there are ways to increase performance with no speed penalty. One of those ways is to apply extra losses to the model during training using modules not executed at test time. This effectively increases feature richness while at no speed penalty.

Thus, we apply a semantic segmentation loss on our feature space using layers that are only evaluated during training. Note that because we construct the ground truth for this loss from instance annotations, this does not strictly capture semantic segmentation (i.e., we do not enforce the standard one class per pixel).
To create predictions during training, we simply attach a 1x1 $\textit{conv}$ layer with $c$ output channels directly to the largest feature map ($P_3$) in our backbone. Since each pixel can be assigned to more than one class, we use sigmoid and $c$ channels instead of softmax and $c+1$. This loss is given a weight of 1 and results in a $+0.4$ mAP boost.

%%%%%%%%%%%% YOLACT++ %%%%%%%%%%%%
    \begin{figure}
    \centering

    \includegraphics[scale=0.35]{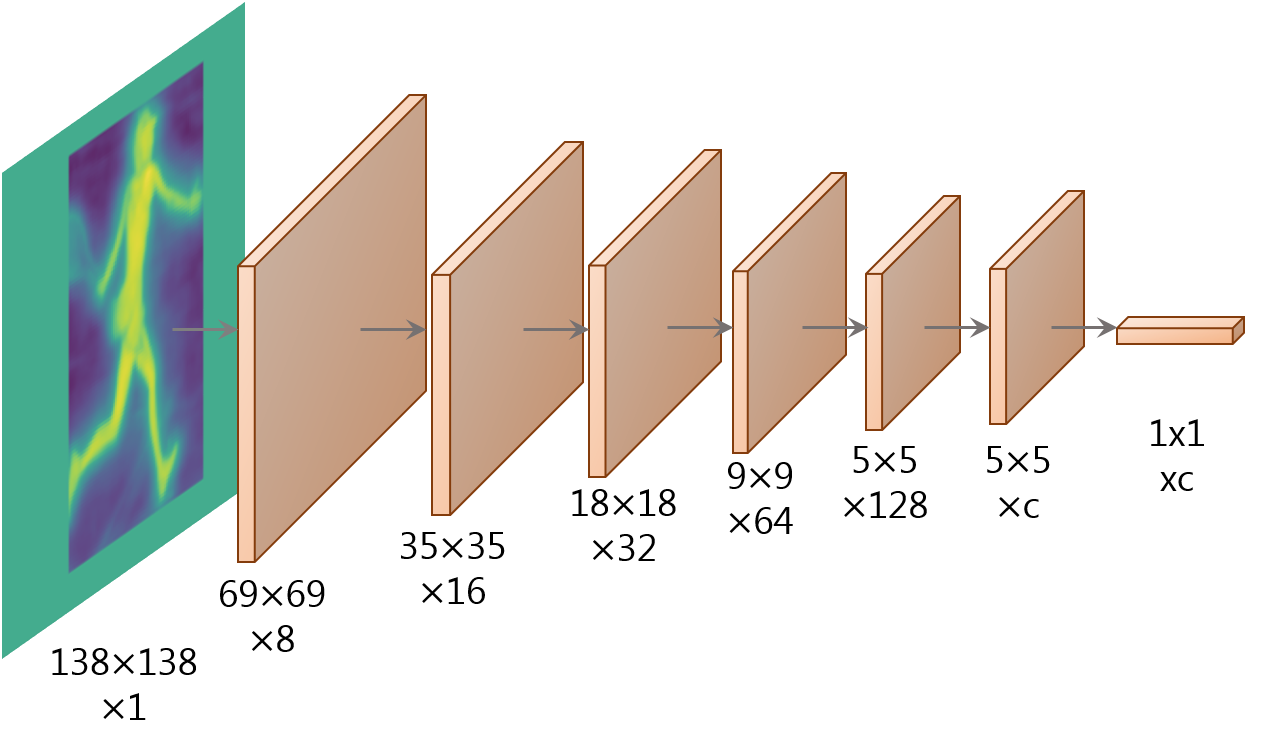}
    % \vspace{-0.06in}
    
    \caption{\inlinesection{Fast Mask Re-scoring Network Architecture} Our mask scoring branch consists of 6 conv layers with ReLU non-linearity and 1 global pooling layer.  Since there is no feature concatenation nor any fc layers, the speed overhead is only $\sim$1 ms.}
    % \vspace{-0.1in}
    \label{fig:mask_score}
\end{figure}
\section{\methodname++}\label{sec:yolact++}
\methodname{}, as introduced thus far, is viable for real-time applications and only consumes $\sim$1500 MB of VRAM even with a ResNet-101 backbone.  We believe these properties make it an attractive model that could be deployed in low-capacity embedded systems. 

We next explore several performance improvements to the original framework, while keeping the real-time demand in mind.  Specifically, we first introduce an efficient and fast mask re-scoring network, which re-ranks the mask predictions according to their mask quality.  We then identify ways to improve the backbone network with deformable convolutions so that our feature sampling aligns better with instances, which results in a better backbone detector and more precise mask prototypes. We finally discuss better choices for the detection anchors to increase recall.% (see Table \ref{tab:yolact++_ablations}).

%\methodname{} can run smoothly to fulfill the demand of real-time applications and only consumes $\sim$1500 MB of VRAM even with the ResNet-101 backbone, which makes it's possible to be deployed into the low-capacity embedded systems. But we don't stop from there because we believe there are still potentials to be explored in the original \methodname{} in terms of the performance. Thus, we take a further investigation into the current failure cases and look up to the recent advances in the community. As the result, with the real-time demand in mind, an efficient and fast mask re-scoring network is proposed and the backbone detector is also upgraded with the deformable convolution with interval and better anchor choice (see Table \ref{tab:yolact++_ablations}).

\subsection{Fast Mask Re-Scoring Network}
As indicated by Mask Scoring R-CNN \cite{huang-msrcnn2018}, there is a discrepancy in the model's classification confidence and the quality of the predicted mask (i.e., higher quality mask segmentations don't necessarily have higher class confidences).  Thus, to better correlate the class confidence with mask quality, Mask Scoring R-CNN adds a new module to Mask R-CNN that learns to regress the predicted mask to its mask IoU with ground-truth.   

Inspired by \cite{huang-msrcnn2018}, we introduce a \emph{fast} mask re-scoring branch, which rescores the predicted masks based on their mask IoU with ground-truth.  Specifically, our Fast Mask Re-Scoring Network is a 6-layer FCN with ReLU non-linearity per conv layer and a final global pooling layer.  It takes as input \methodname{}'s cropped mask prediction (before thresholding) and outputs the mask IoU for each object category. We rescore each mask by taking the product between the predicted mask IoU for the category predicted by our classification head and the corresponding classification confidence (see Figure \ref{fig:mask_score}). 

Our method differs from Mask Scoring R-CNN \cite{huang-msrcnn2018} in the following important ways: (1) Our input is only the mask at the full image size (with zeros outside the predicted box region) whereas their input is the ROI repooled mask concatenated with the feature from the mask prediction branch, and (2) we don't have any $fc$ layers. These make our method \emph{significantly} faster. Specifically, the speed overhead of adding the Fast Mask Re-Scoring branch to \methodname{} is 1.2 ms, which changes the fps from $34.4$ to $33$ for our ResNet-101 model, while the overhead of incorporating Mask Scoring R-CNN's module into \methodname{} is 28 ms (note that the overhead is particularly large for YOLACT as we need to repool features whereas Mask-RCNN could simply reuse ROI features), which would change the fps from $34.4$ to $17.5$. The speed difference mainly comes from MS R-CNN's usage of the ROI align operation, its $fc$ layers, and the feature concatenation in the input.

    \begin{table*}[t]
    \centering
    
    \def\mrcnn{Mask R-CNN \cite{maskrcnn}}
    \def\fcis{FCIS \cite{fcis}}
    \def\panet{PA-Net~\cite{liu-panet2018}}
    \def\msrcnn{MS R-CNN~\cite{huang-msrcnn2018}}
    \def\retinamask{RetinaMask~\cite{fu-retinamask2019}}
    \def\masklab{MaskLab~\cite{chen-masklab2018}}

    \newcommand{\modelname}[1]{\methodname{}-#1}
    
    \begin{smalltable}{l c l cc ccc c ccc cc} \toprule
        Method          && Backbone  &&    FPS    &    Time   &&    AP     & AP$_{50}$ & AP$_{75}$ &&  AP$_{S}$ &  AP$_{M}$ &  AP$_{L}$ \\
        \midrule
        \panet          && R-50-FPN  &&    4.7    &     212.8 &&      36.6 &      58.0 &      39.3 &&      16.3 &      38.1 &      53.1 \\
        \retinamask     && R-101-FPN &&    6.0    &     166.7 &&      34.7 &      55.4 &      36.9 &&      14.3 &      36.7 &      50.5 \\
        \fcis           && R-101-C5  &&    6.6    &     151.5 &&      29.5 &      51.5 &      30.2 &&      8.0  &      31.0 &      49.7 \\
        \mrcnn          && R-101-FPN &&    8.6    &     116.3 &&      35.7 &      58.0 &      37.8 &&      15.5 &      38.1 &      52.4 \\
        \msrcnn         && R-101-FPN &&    8.6    &     116.3 && {\bf 38.3}&      58.8 &      41.5 &&      17.8 &      40.4 &      54.4\\
        \modelname{550} && R-101-FPN && {\bf 33.5}& {\bf 29.8}&&      29.8 &      48.5 &      31.2 &&       9.9 &      31.3 &      47.7  \\
        \midrule
        \modelname{400} && R-101-FPN &&   45.3    &   22.1 &&      24.9 &     42.0 &      25.4 &&       5.0 &      25.3 &      45.0  \\
        \modelname{550} &&  R-50-FPN &&   45.0    &   22.2 &&      28.2 &     46.6 &      29.2 &&       9.2 &      29.3 &      44.8  \\
        \modelname{550} &&  D-53-FPN &&   40.7    &   24.6 &&      28.7 &     46.8 &      30.0 &&       9.5 &      29.6 &      45.5  \\
        \modelname{700} && R-101-FPN &&   23.4    &   42.7 &&      31.2 &     50.6 &      32.8 &&      12.1 &      33.3 &      47.1  \\
        \midrule
        \modelname{550}++ && R-50-FPN &&   33.5    &   29.9 &&      34.1 &     53.3 &      36.2 &&      11.7 &      36.1 &      53.6  \\
        \modelname{550}++ && R-101-FPN &&   27.3    &   36.7 &&      34.6 &     53.8 &      36.9 &&      11.9 &      36.8 &      55.1  \\
        \bottomrule
    \end{smalltable}
    
    % \vspace{-0.05in}
    \caption{\inlinesection{MS COCO \cite{coco} Results} We compare to state-of-the-art methods for mask mAP and speed on COCO {\tt test-dev} and include several ablations of our base model, varying backbone network and image size. We denote the backbone architecture with {\tt network-depth-features}, where {\tt R} and {\tt D} refer to ResNet~\cite{resnet} and DarkNet~\cite{yolov3}, respectively. Our base model, \methodname{}-550 with ResNet-101, is 3.9x faster than the previous fastest approach with competitive mask mAP. Our \methodname{}++-550 model with ResNet-50 has the same speed while improving the performance of the base model by 4.3 mAP. Compared to Mask R-CNN, \methodname{}++-R-50 is 3.9x faster and falls behind by only 1.6 mAP.}
    
    \label{tab:performance}
\end{table*}
%\inlinesection{Mask Performance} We compare our approach to other state-of-the-art methods for mask mAP and speed on COCO {\tt test-dev} and include several ablations of our base model, varying backbone network and image size. We denote the backbone architecture with the nomenclature {\tt network-depth-features}, where {\tt R} and {\tt D} refer to ResNet~\cite{resnet} and DarkNet~\cite{yolov3}, respectively. Our base model, \methodname{}-550 with ResNet-101, is 3.8x faster than the previous fastest approach with competitive mask mAP.
    \begin{table*}[t]
    \centering

    \begin{subfigure}[t]{.36\textwidth}
        \vskip 0pt
        \centering
        \def\mrcnn{Mask R-CNN \cite{maskrcnn}}
        \begin{smalltable}{c r c c c}\toprule
        Method                          &   NMS         &   AP      &  FPS     & Time         \\
        \midrule
        \multirow{2}{*}{\methodname{}}  &   Standard    &{\bf 30.0} & 24.0     & 41.6         \\
                                        &   Fast        & 29.9      &{\bf 33.5}&{\bf 29.8}    \\
        \midrule
        \multirow{2}{*}{Mask R-CNN}     &   Standard    &{\bf 36.1} &  8.6      & 116.0        \\
                                        &   Fast        & 35.8      & {\bf 9.9} &{\bf 101.0}   \\
        \bottomrule
        \end{smalltable}
        \caption{\inlinesection{Fast NMS} Fast NMS performs only slightly worse than standard NMS, while being around 12 ms faster. We also observe a similar trade-off implementing Fast NMS in Mask R-CNN.}
        \label{tab:nms}
    \end{subfigure}
    % \,\,\,
    \qquad
    \begin{subfigure}[t]{.21\textwidth}
        \vskip 0pt
        \centering
        \begin{smalltable}{r c c c}\toprule
            $k$ &   AP &  FPS & Time \\
            \midrule
            8 & 26.8 & {\bf 33.0} & {\bf 30.4} \\
            16 & 27.1 & 32.8 & 30.5 \\
            $\vspace{0pt}^*$32  & 27.7 & 32.4 & 30.9 \\
            64  &{\bf 27.8}& 31.7 & 31.5 \\
            128 & 27.6 & 31.5 & 31.8 \\
            256 & 27.7 & 29.8 & 33.6 \\
            \bottomrule
        \end{smalltable}
        \caption{\inlinesection{Prototypes} Choices for $k$. We choose 32 for its mix of performance and speed.}
        \label{tab:num_proto}
    \end{subfigure}
    % \,\,\,
    \qquad
    \begin{subfigure}[t]{.36\textwidth}
        % \vskip 0.1in
        \vskip 0pt
        \centering
        \begin{smalltable}{l c c c}\toprule
            Method                      &   AP      &   FPS     &   Time    \\
            \midrule
            FCIS w/o Mask Voting        &   27.8    &   9.5     &   105.3   \\
            Mask R-CNN (550 $\times$ 550)              &   {\bf 32.2}&   13.5    &   73.9    \\
            \textit{fc}-mask            &   20.7    &   25.7    &   38.9    \\
            \midrule
            \methodname{}-550 (Ours)    &   29.9    & {\bf 33.5}  & {\bf 29.8}   \\
            \bottomrule
        \end{smalltable}
        \caption{\inlinesection{Accelerated Baselines} We compare to other baseline methods by tuning their speed-accuracy trade-offs. \textit{fc}-mask is our model but with $16 \times 16$ masks produced from an \textit{fc} layer.}
        \label{tab:accelerated_baselines}
    \end{subfigure}
    % \,\,\, 
    % \qquad
    % \begin{subfigure}[t]{.4\textwidth}
    %     \vskip 0.1in
    %     \centering
    %     \begin{smalltable}{l c c c c}\toprule
    %         Method              &   AP$_{mask}$ & AP$_{bbox}$   &   FPS     &   Time    \\
    %         \midrule
    %         \methodname{}-550   &   27.7        &29.8           &32.4       &  30.9   \\
    %         %w/ RetinaNet Head   &   27.9        &31.2           &26.9       &  37.1    \\
    %         w/ 5 Aspect Ratios  &   28.0        &30.1           &{\bf33.2}  &  {\bf30.1}\\
    %         w/ 3 Scales         &   {\bf30.2}   &{\bf32.5}      &30.8       &  32.5    \\
    %         \bottomrule
    %     \end{smalltable}
    %     \caption{\inlinesection{Prediction Head} We compare different design choices of the prediction head by varying the network architecture and adjusting anchor aspect ratios and scales.}
    %     \label{tab:pred_head_ablation}
    % \end{subfigure}

    %\vspace{-0.1in}
    \caption{\inlinesection{Ablations} All models evaluated on COCO \texttt{val2017} using our servers. Models in Table \ref{tab:num_proto} were trained for 400k iterations instead of 800k. Time in milliseconds reported for convenience.}
    % \caption{\inlinesection{Ablations} All models evaluated on COCO \texttt{val2017} using our servers. Models in Table \ref{tab:num_proto} and \ref{tab:pred_head_ablation} were trained for 400k iterations instead of 800k. Time in milliseconds reported for convenience.}
    %\vspace{-0.12in}
    \label{tab:ablations}
\end{table*}

\subsection{Deformable Convolution with Intervals}

Deformable Convolution Networks (DCNs) \cite{deformv1, deformv2} have proven to be effective for object detection, semantic segmentation, and instance segmentation due to its replacement of the rigid grid sampling used in conventional convnets with free-form sampling. We follow the design choice made by DCNv2 \cite{deformv2} and replace the 3x3 convolution layer in each ResNet block with a 3x3 deformable convolution layer for $C_3$ to $C_5$. Note that we do not use the modulated deformable modules because we can't afford the inference time overhead that they introduce. 

Adding deformable convolution layers into the backbone of \methodname{}, leads to a $+1.8$ mask mAP gain with a speed overhead of $8$ ms. We believe the boost is due to: (1) DCN can strengthen the network's capability of handling instances with different scales, rotations, and aspect ratios by aligning to the target instances. (2) \methodname{}, as a single-shot method, does not have a re-sampling process. Thus, a better and more flexible sampling strategy is more critical to \methodname{} than two-stage methods, such as Mask R-CNN because there is no way to recover sub-optimal samplings in our network.  In contrast, the ROI align operation in Mask R-CNN can address this problem to some extent by aligning all objects to a canonical reference region.

Even though the performance boost is fairly decent when directly plugging in the deformable convolution layers following the design choice in \cite{deformv2}, the speed overhead is quite significant as well (see Table \ref{tab:dcn_ablations}). This is because there are 30 layers with deformable convolutions when using ResNet-101. To speed up our ResNet-101 model while maintaining its performance boost, we explore using less deformable convolutions.  Specifically, we try having deformable convolutions in four different configurations: (1) in the last $10$ ResNet blocks, (2) in the last $13$ ResNet blocks, (3) in the last 3 ResNet stages with an interval of $3$ (i.e., skipping two ResNet blocks in between; total 11 deformable layers), and (4) in the last 3 ResNet stages with an interval of $4$ (total 8 deformable layers).  Given the results in Table~\ref{tab:dcn_ablations}, the \textit{DCN (interval=3)} setting is chosen as the final configuration in \methodname++, which cuts down the speed overhead by $5.2$ ms to $2.8$ ms and only has a $0.2$ mAP drop compared to not having an interval.

    \begin{figure*}
    \centering
    \includegraphics[width=0.9\textwidth]{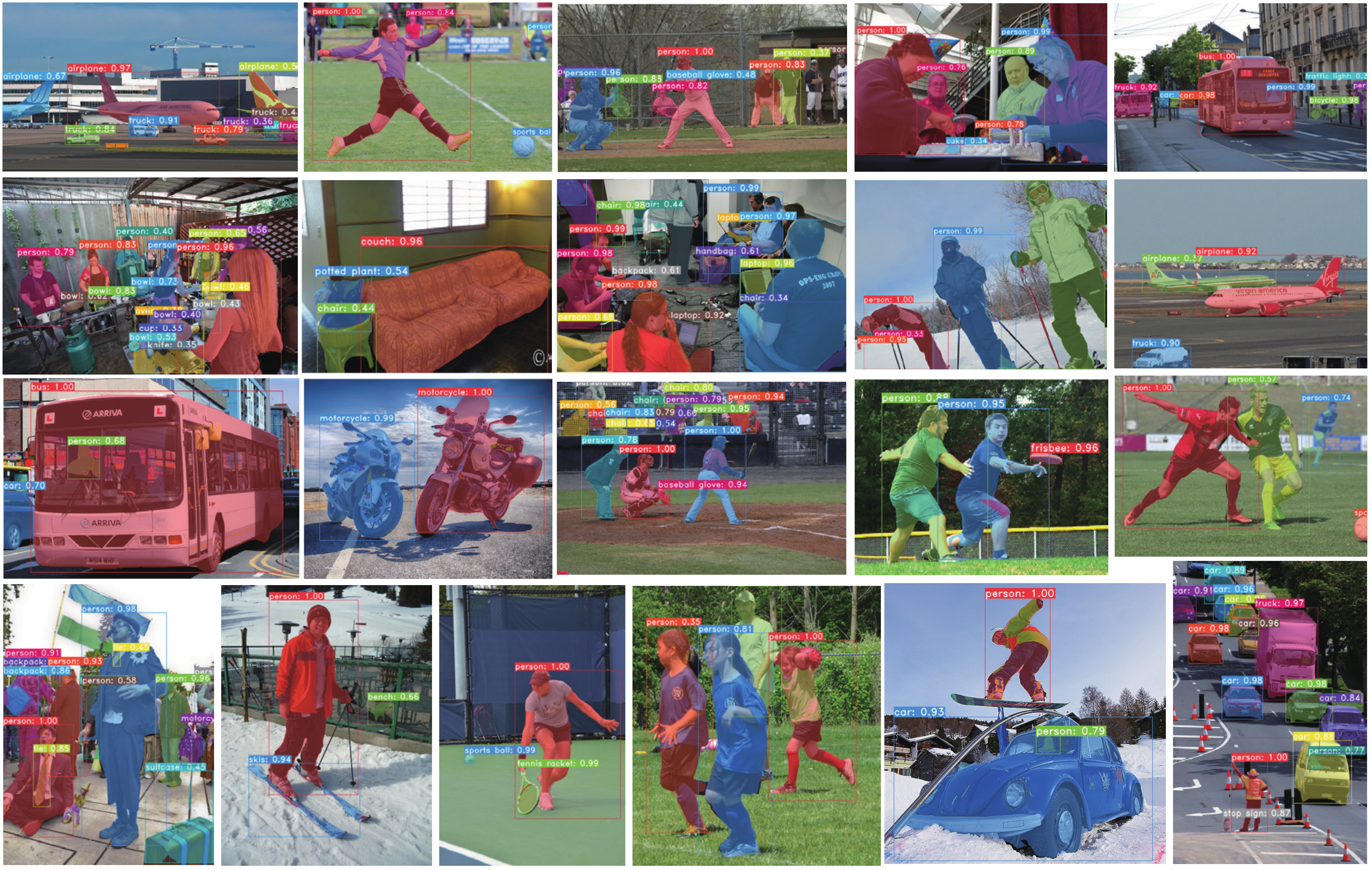}
    % \vspace{-0.12in}
    \caption{\textbf{\methodname} evaluation results on COCO's \texttt{test-dev} set. This base model achieves 29.8 mAP at 33.0 fps. All images have the confidence threshold set to 0.3.} %\reviewonly{Please see the supplementary details for real-time video results.}
    % \vspace{-0.15in}
    \label{fig:qualitative}
\end{figure*}
    
\subsection{Optimized Prediction Head}
Finally, as \methodname{} is based off of an anchor-based backbone detector, choosing the right hyper-parameters for the anchors, such as their scales and aspect ratios, is very important. We therefore revisit our anchor choice and compare with the anchor design of RetinaNet \cite{retinanet} and RetinaMask \cite{fu-retinamask2019}. We try two variations: (1) keeping the scales unchanged while increasing the anchor aspect ratios from $[1, 1/2, 2]$ to $[1, 1/2, 2, 1/3, 3]$, and (2) keeping the aspect ratios unchanged while increasing the scales per FPN level by threefold ([1x, $2^{\frac{1}{3}}$x, $2^{\frac{2}{3}}$x]). The former and latter increases the number of anchors compared to the original configuration of \methodname{} by $\frac{5}{3}$x and 3x, respectively. As shown in Table \ref{tab:pred_head_ablation}, using multi-scale anchors per FPN level (config 2) produces the best speed vs.~performance trade off.

%\chong{As the \methodname{} is based off an anchor-based backbone detector, choosing the right hyper-parameters for the anchors, such as scales and aspect ratios, is quite critical. Thus, we revisit our anchor choice and compare with the anchor design of RetinaNet \cite{retinanet} and RetinaMask \cite{fu-retinamask2019}. Two variations are tried: (1) keep the base scale per FPN level unchanged and expand the anchor aspect ratios from $[1, 1/2, 2]$ to $[1, 1/2, 2, 1/3, 3]$, and (2) keep the base scale and aspect ratios both untouched but for each position while multiple anchor scales per FPN level ($[1, 2^{\frac{1}{3}}, 2^{\frac{2}{3}}]$) are used. We've also tried the RetinaNet prediction head architecture with our original anchor choice. As shown in Table \ref{tab:pred_head_ablation}, using multi-scale anchor per FPN level stands out in terms of speed vs.~performance trade off.}

%%%%%%%%%%%% RESULTS %%%%%%%%%%%%
% \vspace{-0.05in}
\section{Results} \label{sec:results}
% \vspace{-0.05in}

%We report results on MS COCO's instance segmentation task \cite{coco} using the standard metrics for the task. We train on {\tt train2017} and evaluate on {\tt val2017} and {\tt test-dev}. 
We report instance segmentation results on MS COCO \cite{coco} and Pascal 2012 SBD~\cite{sbd} using the standard metrics. For MS COCO, we train on {\tt train2017} and evaluate on {\tt val2017} and {\tt test-dev}.  We also report box detection results on MS COCO.

% \vspace{-0.01in}
\subsection{Implementation Details}
We train all models with batch size 8 \textit{on one GPU} using ImageNet \cite{imagenet} pretrained weights. We find that this is a sufficient batch size to use batch norm, so we leave the pretrained batch norm unfrozen but do not add any extra \textit{bn} layers. We train with SGD for 800k iterations starting at an initial learning rate of $10^{-3}$ and divide by 10 at iterations 280k, 600k, 700k, and 750k, using a weight decay of $5{\times}10^{-4}$, a momentum of 0.9, and all data augmentations used in SSD \cite{ssd}. For Pascal, we train for 120k iterations and divide the learning rate at 60k and 100k. We also multiply the anchor scales by $4/3$, as objects tend to be larger. Training takes 4-6 days (depending on config) on one Titan Xp for COCO and less than 1 day on Pascal.

\subsection{Mask Results}
We first compare \methodname{} to state-of-the art methods on COCO's {\tt test-dev} set in Table \ref{tab:performance}. Because our main goal is speed, we compare against other single model results with no test-time augmentations. We report all speeds computed on a single Titan Xp, so some listed speeds may be faster than in the original paper.
    
\begin{figure*}[t!]
    \centering
    \includegraphics[width=0.97\textwidth]{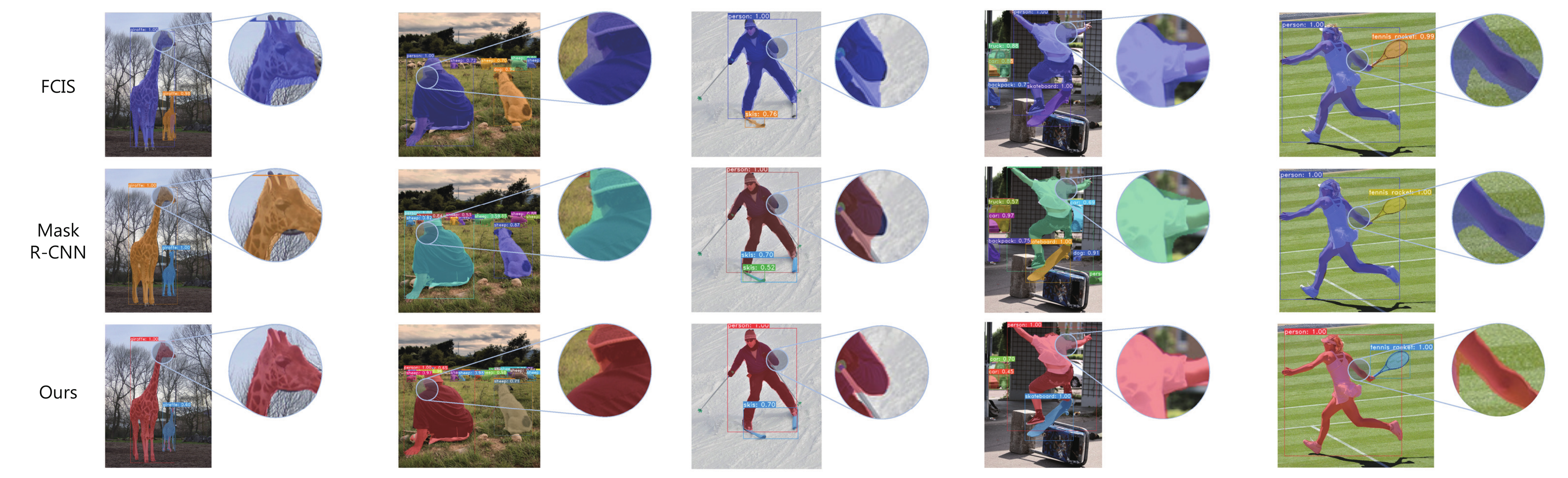}
    % \vspace{-0.13in}
    \caption{\inlinesection{Mask Quality} Our masks are typically higher quality than those of Mask R-CNN \cite{maskrcnn} and FCIS \cite{fcis} because of the larger mask size and lack of feature repooling.}
    % \vspace{-0.13in}
    \label{fig:mask_quality}
\end{figure*}

\methodname{}-550 offers competitive instance segmentation performance while at 3.8x the speed of the previous fastest instance segmentation method on COCO. We also note an interesting difference in where the performance of our method lies compared to others. Supporting our qualitative findings in Figure~\ref{fig:mask_quality}, the gap between \methodname{}-550 and Mask R-CNN at the 50\% overlap threshold is 9.5 AP, while it's 6.6 at the 75\% IoU threshold. This is different from the performance of FCIS, for instance, compared to Mask R-CNN where the gap is consistent (AP values of 7.5 and 7.6 respectively). Furthermore, at the highest (95\%) IoU threshold, we outperform Mask R-CNN with 1.6 vs.~1.3 AP.

%of our model 
We also report numbers for alternate model configurations in Table~\ref{tab:performance}. In addition to our base $550\times 550$ image size model, we train $400 \times 400$ (\methodname{}-400) and $700 \times 700$ (\methodname{}-700) models, adjusting the anchor scales accordingly ($s_x = s_{550} / 550 * x$). Lowering the image size results in a large decrease in performance, demonstrating that instance segmentation naturally demands larger images. Then, raising the image size decreases speed significantly but also increases performance, as expected.  In addition to our base backbone of ResNet-101 \cite{resnet}, we also test ResNet-50 and DarkNet-53 \cite{yolov3} to obtain even faster results. If higher speeds are preferable we suggest using ResNet-50 or DarkNet-53 instead of lowering the image size, as these configurations perform much better than \methodname{}-400, while only being slightly slower.
%for these models

    \begin{table}[t]
    \centering

    \newcommand{\modelname}[1]{\methodname{}-#1}

    \begin{smalltable}{l c c c c} 
        \toprule
        Method              &   AP$_{mask}$ & AP$_{bbox}$   &   FPS     &   Time    \\
        \midrule
        \methodname{}-550   &   27.7        &29.8           &32.4       &  30.9   \\
        %w/ RetinaNet Head   &   27.9        &31.2           &26.9       &  37.1    \\
        w/ 5 Aspect Ratios  &   28.0        &30.1           &{\bf33.2}  &  {\bf30.1}\\
        w/ 3 Scales         &   {\bf30.2}   &{\bf32.5}      &31.2       &  32.1    \\
        \bottomrule
    \end{smalltable}
    
    %\vspace{-0.05in}
    \caption{\inlinesection{Different Anchor Choices of Prediction Head} We compare different anchor aspect ratios and scales. All models were trained for 400k iterations. Results on MS COCO {\tt val2017}.}
    \label{tab:pred_head_ablation}
\end{table}
%design choices of the prediction head by adjusting 
    
The bottom two rows in Table~\ref{tab:performance} show the results of our \methodname++ model with ResNet-50 and ResNet-101 backbones. With the proposed enhancements, \methodname++ obtains a huge performance boost over \methodname{} ($5.9$ mAP for the ResNet-50 model and $4.8$ mAP for the ResNet-101 model) while maintaining high speed. In particular, our \methodname++-ResNet-50 model runs at a real-time speed of $33.5$ fps, which is 3.9x faster than Mask R-CNN, while its instance segmentation accuracy only falls behind by $1.6$ mAP.

Finally, we also train and evaluate our YOLACT ResNet-50 model on Pascal 2012 SBD in Table~\ref{tab:pascal}. \methodname{} clearly outperforms popular approaches that report SBD performance, while also being significantly faster.

%\chong{Finally, we show the numbers from our \methodname++ model with ResNet-50 and ResNet-101 backbones. With the proposed enhancements, our model witnesses a huge performance boost ($5.9$ mAP for the ResNet-50 model and $4.8$ mAP for the ResNet-101 model) while still keeping running at high speed. In particular, our \methodname++-ResNet-50 model runs at a real-time speed of $33.5$ fps, which is 3.9x faster than Mask R-CNN, while the performance only falls behind by $1.6$ mAP.}

    \begin{table}
    \centering
    
    \def\mnc{MNC \cite{mnc}}
    \def\fcis{FCIS \cite{fcis}}

    \newcommand{\modelname}[1]{\methodname{}-#1}
    
    \begin{smalltable}{l c l cc ccc c ccc cc} \toprule
        Method          & Backbone  &    FPS    &    Time   & $\text{mAP}^r_{50}$ & $\text{mAP}^r_{70}$ \\
        \midrule
        \mnc            &   VGG-16  &    2.8    &      360 &        63.5         &         41.5        \\
        \fcis           & R-101-C5  &    9.6    &      104 &        65.7         &         52.1        \\
        \modelname{550} & R-50-FPN  &{\bf  47.6}&{\bf  21.0}&      {\bf 72.3}     &      {\bf 56.2}     \\
        \bottomrule
    \end{smalltable}
    
    % \vspace{-0.05in}
    \caption{\inlinesection{Pascal 2012 SBD \cite{sbd} Results} Timing for FCIS redone on a Titan Xp for fairness. Since Pascal has fewer and easier detections than COCO, YOLACT does much better than previous methods. Note that COCO and Pascal FPS are not comparable because Pascal has fewer classes.}
    
    \label{tab:pascal}
    % \vspace{-0.1in}
\end{table}

%\caption{\inlinesection{Pascal SBD Performance} YOLACT mask performance on Pascal VOC 2012 \cite{pascalvoc} with SBD annotations \cite{sbd}. Because there are fewer detections in Pascal VOC, and the detections are easier to localize than with COCO, YOLACT does much better on Pascal than previous methods applied to the dataset. All timing was done on our server with one Titan Xp, since the older methods used worse hardware. Note that Pascal FPS is not comparable to COCO FPS because Pascal has fewer classes and thus NMS takes less time to evaluate.}
% These images were evaluated on the 29.2 mAP version of FCIS and 35.7 mAP version of Mask R-CNN.
    \begin{figure*}
    \centering
    \includegraphics[width=0.9\textwidth]{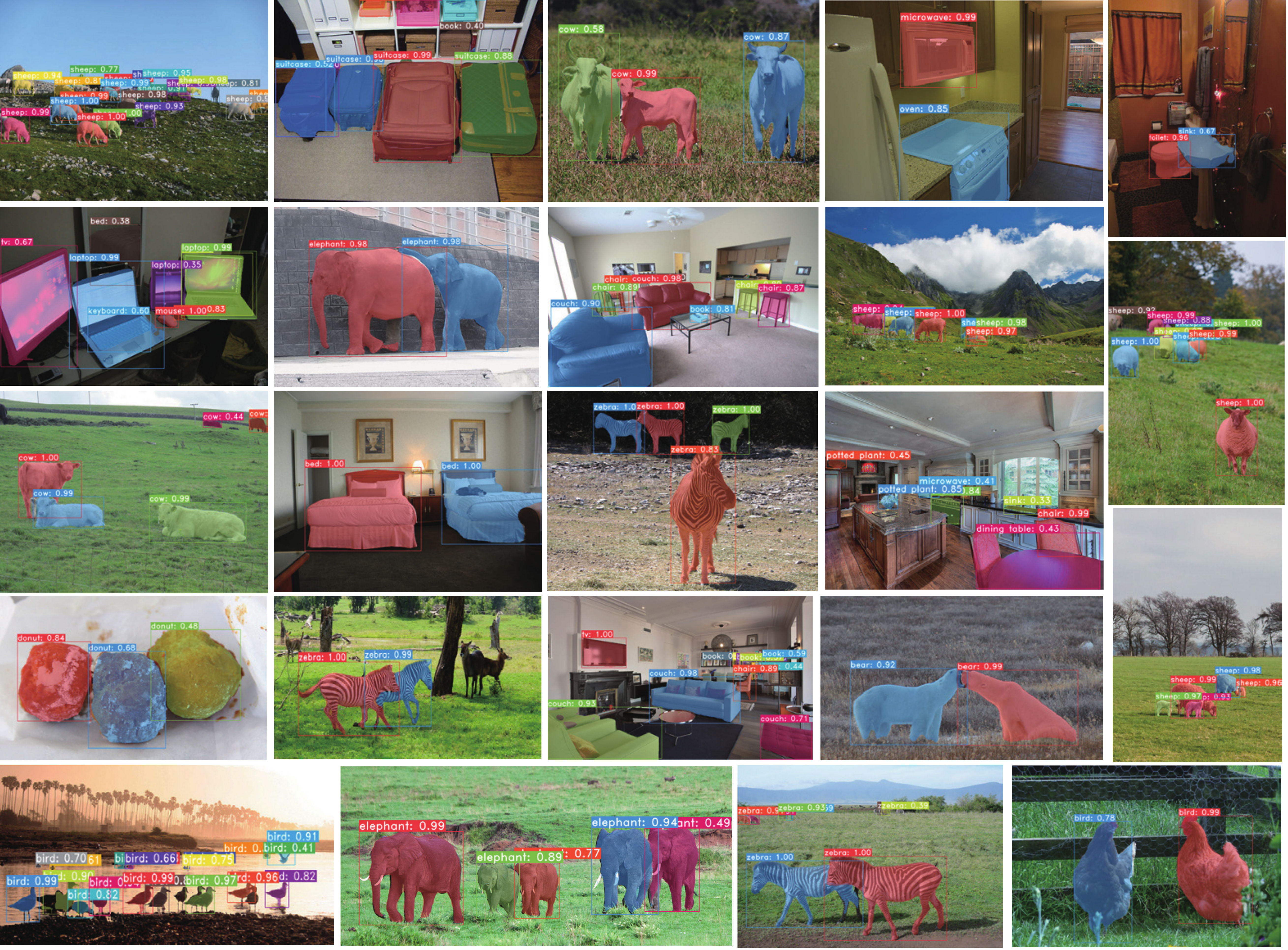}
    % \vspace{-0.05in}
    \caption{\textbf{More \methodname{}} evaluation results on COCO's \texttt{test-dev} set with the same parameters as before. To further support that \methodname{} implicitly localizes instances, we select examples with adjacent instances of the same class.}
    % \vspace{-0.1in}
    \label{fig:more_qualitative}
\end{figure*}
    \begin{table*}
    \centering
    \newcommand{\yolo}[1]{YOLOv3-#1 \cite{yolov3}}
    \newcommand{\modelname}[1]{\methodname{}-#1}
    
    \begin{smalltable}{l c l cc ccc c ccc cc} \toprule
        Method          && Backbone  &&    FPS                  & Time              &&  AP              & AP$_{50}$ & AP$_{75}$ &&  AP$_{S}$ &  AP$_{M}$ &  AP$_{L}$    \\
        \midrule
        \yolo{320}      && D-53      &&    \textbf{71.17}       & \textbf{14.05}    &&  28.2            & 51.5       & \ph       &&  \ph      &  \ph      &  \ph         \\
        \modelname{400} && R-101-FPN &&    55.43                & 18.04             &&  \textbf{28.4}   & 48.6      & 29.5      &&  10.7     &  28.9     &  43.1        \\
    	\midrule
    	\modelname{550} && R-50-FPN  &&    \textbf{59.30}       & \textbf{16.86}    &&  30.3            & 50.8      & 31.9      &&  14.0     &  31.2     &  43.0        \\
    	\yolo{416}      && D-53      &&    54.47                & 18.36             &&  \textbf{31.0}   & 55.3       & \ph       &&  \ph      &  \ph      &  \ph         \\
        \modelname{550} && D-53-FPN  &&    52.99                & 18.87             &&  \textbf{31.0}   & 51.1      & 32.9      &&  14.4     &  31.8     &  43.7        \\
        % \modelname{550}++ && R-50-FPN &&  -&- &&-&-&-&&-&-&-\\
    	\midrule
        \modelname{550} && R-101-FPN &&    \textbf{41.14}       & \textbf{24.31}    &&  32.3            & 53.0      & 34.3      &&  14.9     &  33.8     &  45.6        \\
    	\yolo{608}      && D-53      &&    30.54                & 32.74             &&  33.0            & 57.9      & 34.4      &&  18.3     &  35.4     &  41.9        \\
    	\modelname{700} && R-101-FPN &&    29.61                & 33.77             &&  \textbf{33.7}   & 54.3      & 35.9      &&  16.8     &  35.6     &  45.7        \\
    % 	\modelname{550}++ && R-101-FPN &&  -&- &&-&-&-&&-&-&-\\
        \bottomrule
    \end{smalltable}
    \caption{\textbf{Box Performance} on COCO's \texttt{test-dev} set. For our method, timing is done without evaluating the mask branch. Both methods were timed on the same machine (using one Titan Xp). In each subgroup, we compare similar performing versions of our model to a corresponding YOLOv3 model. YOLOv3 doesn't report all metrics for the 320 and 416 versions. }
    % \caption{\textbf{\textit{Box} Performance} on COCO's \texttt{test-dev} set. For our method, timing is done without evaluating the mask branch \chong{and fast mask re-scoring branch}. Both methods were timed on the same machine (using one Titan Xp). In each subgroup, we compare similar performing versions of our model to a corresponding YOLOv3 model. YOLOv3 doesn't report all metrics for the 320 and 416 versions. }
    \label{tab:detection}
\end{table*}

\subsection{Mask Quality} %\label{sec:mask_quality}
Because we produce a final mask of size $138 \times 138$, and because we create masks directly from the original features (with no repooling to transform and potentially misalign the features), our masks for large objects are noticeably higher quality than those of Mask R-CNN \cite{maskrcnn} and FCIS \cite{fcis}. For instance, in Figure~\ref{fig:mask_quality}, \methodname{} produces a mask that cleanly follows the boundary of the arm, whereas both FCIS and Mask R-CNN have more noise. Moreover, despite being 5.9 mAP worse overall, at the 95\% IoU threshold, our base model achieves 1.6 AP while Mask R-CNN obtains 1.3. This indicates that repooling does result in a quantifiable decrease in mask quality.
%step

%Although we only train our model using static images and do not apply any temporal smoothing, we find that the model produces more temporally stable masks on videos than Mask R-CNN, whose masks jitter across frames even when objects are completely stationary.
\subsection{Temporal Stability}
Although we only train using static images and do not apply any temporal smoothing, we find that our model produces more temporally stable masks on videos than Mask R-CNN, whose masks jitter across frames even when objects are stationary. We believe our masks are more stable in part because they are higher quality (thus there is less room for error between frames), but mostly because our model is one-stage. Masks produced in two-stage methods are highly dependent on their region proposals in the first stage. In contrast for our method, even if the model predicts different boxes across frames, the prototypes are not affected, yielding much more temporally stable masks. See \url{https://www.youtube.com/watch?v=Im-bqiWQ5nE} for a comparison between YOLACT Base and Mask R-CNN. %\reviewonly{Please see the supplementary details for real-time video results.}
%temporally

% \subsection{Implementation Details}
% We train all models with batch size 8 \textit{on one GPU} using ImageNet \cite{imagenet} pretrained weights. We find that this is a sufficient batch size to use batch norm, so we leave the pretrained batch norm unfrozen but do not add any extra \textit{bn} layers. We train with SGD for 800k iterations starting at an initial learning rate of $10^{-3}$ and divide by 10 at iterations 280k, 600k, 700k, and 750k, using a weight decay of $5{\times}10^{-4}$ and a momentum of 0.9. We also train with all data augmentations used in SSD \cite{ssd}. \daniel{Training takes 4-6 days (depending on config) on one Titan Xp.}

%%%%%%%%%%%% BIBLIOGRAPHY %%%%%%%%%%%%

% if have a single appendix:
%\appendix[Proof of the Zonklar Equations]
% or
%\appendix  % for no appendix heading
% do not use \section anymore after \appendix, only \section*
% is possibly needed

% use appendices with more than one appendix
% then use \section to start each appendix
% you must declare a \section before using any
% \subsection or using \label (\appendices by itself
% starts a section numbered zero.)
%

% \appendices
\subsection{More Qualitative Results}
    Figure~\ref{fig:qualitative} shows many examples of adjacent people and vehicles, but not many for other classes. To further support that \methodname{} is not just doing semantic segmentation, we include many more qualitative results for images with adjacent instances of the same class in Figure~\ref{fig:more_qualitative}.
    
    For instance, in an image with two elephants (Figure~\ref{fig:more_qualitative} row 2, col 2), despite the fact that two instance boxes are overlapping with each other, their masks are clearly separating the instances. This is also clearly manifested in the examples of zebras (row 4, col 2) and birds (row 5, col 1). 
    
    Note that for some of these images, the box doesn't exactly crop off the mask. This is because for speed reasons (and because the model was trained in this way), we crop the mask at the prototype resolution (so one fourth the image resolution) with 1px of padding in each direction. On the other hand, the corresponding box is displayed at the original image resolution with no padding.

\subsection{Box Results}
    Since \methodname{} produces boxes in addition to masks, we can also compare its object detection performance to other real-time object detection methods. Moreover, while our \textit{mask performance} is real-time, we don't need to produce masks to run \methodname{} as an object detector. Thus, \methodname{} is faster when run to produce boxes than when run to produce instance segmentations. 
    
    In Table~\ref{tab:detection}, we compare our performance and speed to various skews of YOLOv3 \cite{yolov3}.  We are able to achieve similar detection results to YOLOv3 at similar speeds, while not employing any of the additional improvements in YOLOv2 and YOLOv3 like multi-scale training, optimized anchor boxes, cell-based regression encoding, and objectness score. Because the improvements to our detection performance in our observation come mostly from using FPN and training with masks (both of which are orthogonal to the improvements that YOLO makes), it is likely that we can combine YOLO and \methodname{} to create an even better detector.

    Moreover, these detection results show that our mask branch takes \emph{only 6 ms} in total to evaluate, which demonstrates how minimal our mask computation is.

    \begin{table}[t]
    \centering

    \newcommand{\modelname}[1]{\methodname{}-#1}

    \begin{smalltable}{l c l cc ccc c ccc cc} 
        \toprule
        Method                      &    FPS    &    Time   & AP        \\
        \midrule
        \vspace{0.01in}
        \modelname{550} (R-101-FPN) &   {\bf 33.5}    &   {\bf 29.8}    & 29.9     \\
        \vspace{0.01in}
        + more anchors               &   30.8    &   32.5    & 31.7      \\
        \vspace{0.01in}
        + deform convs (interval=3)  &   28.3    &   35.3    & 33.3      \\
        \vspace{0.01in}
        + fast mask re-scoring       &   27.3    &   36.7    & {\bf 34.4}      \\
        \midrule
        \vspace{0.01in}
        \modelname{550} (R-50-FPN)  &   {\bf 45.0}    &   {\bf 22.2}    & 28.5     \\
        \vspace{0.01in}
        + more anchors               &   40.2    &   24.9    & 29.9     \\
        \vspace{0.01in}
        + deform convs               &   34.7    &   28.8    & 32.7      \\
        \vspace{0.01in}
        + fast mask re-scoring       &   33.5    &   29.9    & {\bf 33.7}      \\
        \bottomrule
    \end{smalltable}
    
   % \vspace{-0.05in}
    \caption{\inlinesection{\methodname++ Improvements} Contribution to instance segmentation accuracy and speed overhead of each component of \methodname++.  Results on MS COCO {\tt val2017}.}
    
    \label{tab:yolact++_ablations}
    %\vspace{-0.1in}
\end{table}

%\chong{The demonstration of how each component of \methodname++ contributes to the performance on MS COCO {\tt val2017} and its corresponding speed overhead.}
    \begin{table}[t]
    \centering

    \newcommand{\modelname}[1]{\methodname{}-#1}

    \begin{smalltable}{l c l cc ccc c ccc cc} 
        \toprule
        Method                      &    FPS        &    Time     & AP        \\
        \midrule
        w/o DCN                 &   {\bf 30.8}       &   {\bf 32.5}       & 31.7     \\
        w/ DCN                    &   24.7       &   40.5       & {\bf 33.5}      \\
        w/ DCN (interval=3)       &   28.3       &   35.3       & 33.3      \\
        w/ DCN (interval=4)       &   29.2       &   34.3       & 33.0      \\
        w/ DCN (last 10 layers)   &   29.0       &   34.5       & 33.0      \\
        w/ DCN (last 13 layers)   &   28.0       &   35.8       & 33.0      \\
        \bottomrule
    \end{smalltable}
    
    %\vspace{-0.05in}
    \caption{\inlinesection{Different Choices of Using Deformable Convolution Layers} The speed vs.~performance trade off of different design choices when applying deformable convolutions~\cite{deformv2} in \methodname{}. Results on MS COCO {\tt val2017}. Note that in these results, the backbone is ResNet-101 with the 3-scale anchor choice in the prediction head.}
    
    \label{tab:dcn_ablations}
    %\vspace{-0.1in}
\end{table}

\subsection{\methodname{}++ Improvements}
    % We already stated the below in Sec 7.2
    %\chong{In Table~\ref{tab:performance}, we show the numbers from our \methodname++ model with ResNet-50 and ResNet-101 backbones. With the proposed enhancements, our model witnesses a huge performance boost ($5.9$ mAP for the ResNet-50 model and $4.8$ mAP for the ResNet-101 model) while still keeping running at high speed. In particular, our \methodname++-ResNet-50 model runs at a real-time speed of $33.5$ fps, which is 3.9x faster than Mask R-CNN, while the performance only falls behind by $1.6$ mAP.}

    \begin{figure*}[t]
    \centering
    \includegraphics[width=0.97\textwidth]{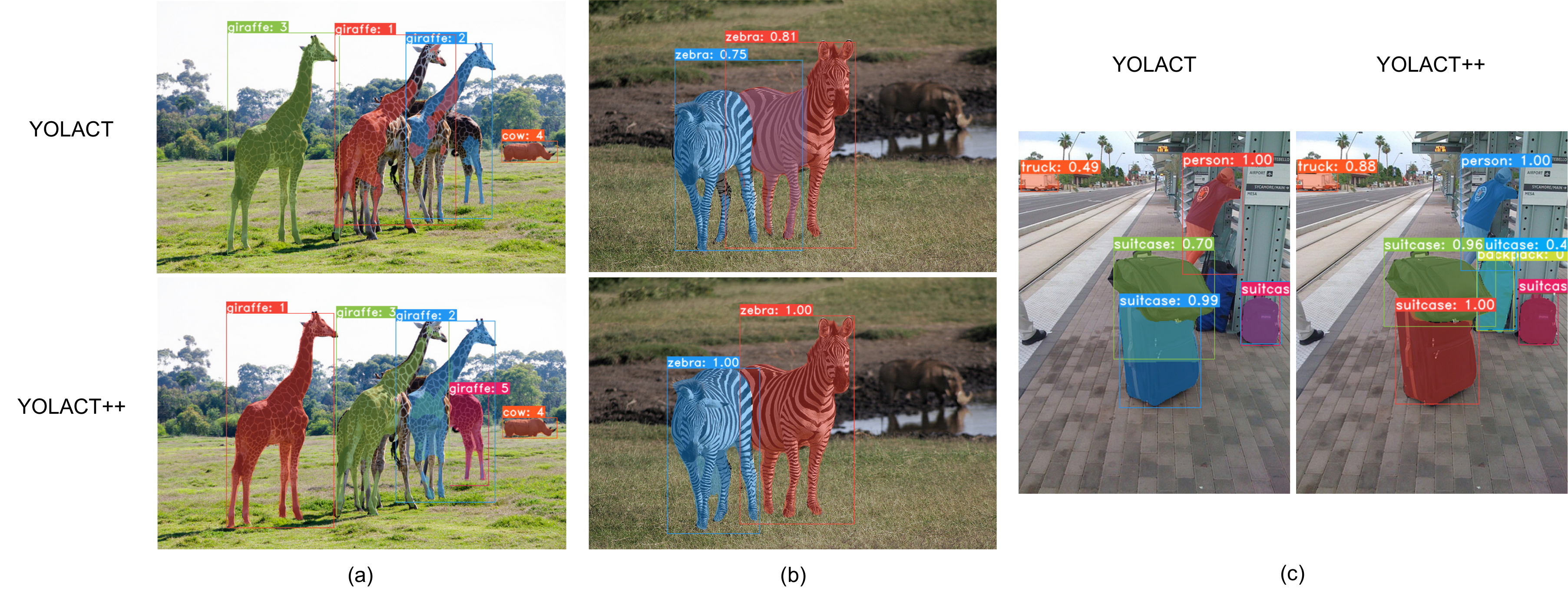}
    \caption{\inlinesection{YOLACT vs. YOLACT++} (a) shows the rank of each detection in the image. As YOLACT++ has a fast mask re-scoring branch, its detections with better masks are ranked higher than those of YOLACT (see the leftmost giraffe). Since YOLACT++ is equipped with deformable convolutions in the backbone and has a better anchor design, the box recall, mask quality, and classification confidence are all increased. Specifically, (b) shows that both the box prediction and instance segmentation mask of the left zebra is more precise. (c) shows increased detection recall and improved class confidence scores.}
    \label{fig:yolact++_comparison}
\end{figure*}

%\chong{(a) shows the rank of each detection. As YOLACT++ has a fast mask re-scoring branch, the detections with better masks are ranked higher than YOLACT (See the giraffe on the far left). (b) and (c) shows the class confidence of each detection. Since YOLACT++ is equipped with DCN in the backbone and better anchor choice, the box recall, mask quality and classification confidence are all increased. Specically, (b) shows not only the box prediction of the left zebra is more precise, but also in terms of masks, the instance identity is more accurate (no leakage on the box overlapping region). (c) clearly shows the increased detection recall and the class confidence is also improved.}
    \begin{table*}[h!]
    \centering

    \newcommand{\modelname}[1]{\methodname{}-#1}

    \begin{smalltable}{l rr c rr c rr c rr}\toprule
	Model & \multicolumn{2}{c}{YOLACT Base}	&&		\multicolumn{2}{c}{YOLACT Resnet50}	&&		\multicolumn{2}{c}{YOLACT++ Base}	&&		\multicolumn{2}{c}{YOLACT++ Resnet50} \\
	   \cmidrule(lr){2-3}
    \cmidrule(lr){5-6}
    \cmidrule(lr){8-9}
    \cmidrule(lr){11-12}
	Component &ms	&\%		&&ms	&\%		&&ms	&\%		&&ms	&\% \\\midrule
    Backbone	&64.66	&63.43\%&		&32.85	&47.53\%&&		70.37&	63.62\%	&&	44.07&	52.37\%\\
    FPN	    &4.39	&4.31\%	&&	5.20	&7.53\%	&&	4.12&	3.72\%	&&	4.12&	4.89\%\\
    Protonet	&5.08	&4.98\%&	&	5.03&	7.28\%&	&	4.63&	4.19\%&&		4.58&	5.44\%\\
    Prediction Heads	&10.72	&10.52\%&&		10.58&	15.3\%&	&	11.94&	10.8\%	&&	11.81&	14.03\%\\
    Fast NMS	& 6.21	&6.09\%	&&	5.87&	8.5\%&&		7.12&	6.44\%	&&	6.96&	8.28\%\\
    Postprocessing	&7.47&	7.33\%	&&	6.16&	8.91\%	&&	5.97&	5.40\%	&&	6.14&	7.29\%\\
    Mask Rescoring	&$\cdot$&$\cdot$&&$\cdot$&$\cdot$&& 2.69&	2.43\%&&		2.72&	3.23\%\\
    Other			&3.41&	3.35\%&&		3.42&	4.95\%&&						3.77&	3.40\%&&		3.76&	4.47\%\\\midrule
    Total	& 101.95&	100\%&	&	69.11&	100\%&	&	110.60&	100\%&&		84.15&	100\%\\
    \bottomrule
    \end{smalltable}
    %Mask IoU Net
    %\vspace{-0.05in}
    \caption{\inlinesection{Timing Breakdown} The time taken for each stage of the model. Note that in order to properly time each portion of the model, we have to disable GPU parallelization (i.e., with \texttt{CUDA\_LAUNCH\_BLOCKING=1}), which means that the times shown here are much higher than what is typical of the model. The fact that this sequential execution of our model is 3 times slower than normal also shows how well our method exploits parallelization.}
    
    \label{tab:timing}
    %\vspace{-0.1in}
\end{table*}
    
     Table~\ref{tab:yolact++_ablations} shows the contribution of each new component in our \methodname{}++ model. The optimized anchor choice directly improves the recall of box prediction and boosts our backbone detector. The deformable convolutions help with better feature sampling by aligning the sampling positions with the instances of interest and better handles changes in scale, rotation, and aspect ratio. Importantly, with our exploration of using less deformable convolution layers, we can cut down their speed overhead significantly (from 8 ms to 2.8 ms) while keeping the performance almost the same (only 0.2 mAP drop) as compared to the original configuration proposed in \cite{deformv2}; see Table~\ref{tab:dcn_ablations}. With these two upgrades for object detection, \methodname{}++  suffers less from localization failure and has finer mask predictions, as shown in Figure \ref{fig:yolact++_comparison}b, c, which together result in 3.4 mAP and 4.2 mAP boost for ResNet-101 and ResNet-50, respectively. In addition, the proposed fast mask re-scoring network re-ranks the mask predictions with the IoU based mask scores instead of solely relying on classification confidence. As a result, the under-estimated masks (masks with good quality but with low classification confidence) and over-estimated masks (masks with bad quality but with high classification confidence) are put into a more proper ranking as shown in Figure \ref{fig:yolact++_comparison}a. Our mask re-scoring method is also fast. Compared to incorporating MS R-CNN into \methodname{}, it is 26.8 ms faster yet can still improve \methodname{} by 1 mAP.
     
    %\chong{Table~\ref{tab:yolact++_ablations} shows how much does each new component contribute. The optimized anchor choice directly improves the recall of box prediction and boosts our backbone detector. The DCN helps with better feature sampling by aligning the sampling positions with the instances of interest and addresses the multiple scales, rotations, and aspect ratios problem in the same time. Besides, with our exploration in using less DCN layers, we cut down the speed overhead introduced by DCN significantly (from 8 ms to 2.8 ms) while keeping the performance almost the same (only 0.2 mAP drop). See Table~\ref{tab:dcn_ablations}. With these two upgrades for object detection, the \methodname{}++ model suffers less from the localization failure and has finer mask predictions as shown in Figure \ref{}, which results in 3.4 mAP and 4.2 mAP boost for ResNet-101 and ResNet-50 seperately. In addition, the proposed fast mask re-scoring network re-ranks the mask predictions with the dedicated mask scores in stead of solely relying on classification confidence. As a result, the over-estimated masks (masks with good quality but have low classification confidences) and under-estimated masks (masks with bad quality but have high classification confidences) are put into more proper rankings as shown in Figure \ref{}. Our mask re-scoring method is also fast and efficient. Compared to incorporating MS R-CNN into \methodname{}, ours is 26.8 ms faster and still can improve the \methodname{} by 1 mAP.}

\subsection{Timing Breakdown} In Table~\ref{tab:timing}, we demonstrate the time taken for each part of our method with asynchronous GPU execution disabled (i.e., \texttt{CUDA\_LAUNCH\_BLOCKING=1}) in order to properly time each part. Note that because this is timed with parallelism turned off, the total time is much higher than the original model, and thus the time of each component has to be considered individually. The fact that our model is 3 times faster with parallelism turned on also demonstrates how effective our method is at exploiting parallel computation.

%%%%%%%%%%%% DISCUSSION %%%%%%%%%%%%
% \vspace{-0.05in}
\section{Discussion}
% \vspace{-0.05in}

Despite our masks being higher quality and having nice properties like temporal stability, we fall a bit behind state-of-the-art instance segmentation methods in overall performance, albeit while being much faster. Most errors are caused by mistakes in the detector: misclassification, box misalignment, etc. However, we have identified two typical errors caused by \methodname{}'s mask generation algorithm.
%simply 

% \vspace{-0.05in}
\paragraph{Localization Failure}
If there are too many objects in one spot in a scene, the network can fail to localize each object in its own prototype. In these cases, it will output something closer to a foreground mask than an instance segmentation for some objects in the group; e.g., in the first image in Figure~\ref{fig:qualitative} (row 1 column 1), the blue truck under the red airplane is not properly localized.
%An example of this can be seen in the first image in Figure~\ref{fig:qualitative} (row 1 column 1), where the blue truck under the red airplane is not properly localized.

Our \methodname++ model addresses this problem to some degree by introducing more anchors covering more scales and applying deformable convolutions in the backbone for better feature sampling. For example, there are higher confidence and more accurate box detections in Figure \ref{fig:yolact++_comparison}c using \methodname{}++.

% \vspace{-0.05in}
\paragraph{Leakage}
Our network leverages the fact that masks are cropped after assembly, and makes no attempt to suppress noise outside of the cropped region. This works fine when the bounding box is accurate, but when it is not, that noise can creep into the instance mask, creating some ``leakage'' from outside the cropped region. This can also happen when two instances are far away from each other, because the network has learned that it doesn't need to localize far away instances---the cropping will take care of it. However, if the predicted bounding box is too big, the mask will include some of the far away instance's mask as well. For instance, Figure~\ref{fig:qualitative} (row 2 column 4) exhibits this leakage because the mask branch deems the three skiers to be far enough away to not have to separate them.
 
%\colorsout[\daniel]{These issues could potentially be mitigated with a mask error down-weighting scheme like in MS R-CNN \cite{huang-msrcnn2018}, where masks exhibiting these errors could be ignored. However, we leave this for future works to address.}
Our \methodname++ model partially mitigates these issues with a light-weight mask error down-weighting scheme, where masks exhibiting these errors will be ignored or ranked lower than higher quality masks. In Figure \ref{fig:yolact++_comparison}a, the leftmost giraffe's mask has the best quality and with mask re-scoring, it is ranked highest with \methodname{}++ whereas with \methodname{} it is ranked 3rd among all detections in the image. %despite its class confidence might not be higher than the rest.

%\subsection
% \vspace{-0.05in}
\paragraph{Understanding the AP Gap}
However, localization failure and leakage alone are not enough to fully explain the gap between YOLACT's base model and, say, Mask R-CNN. Indeed, if we ignore all mask-related errors and replace the predicted masks with the ground-truth, our mask mAP only improves from 33.7 to 35.1 (given 34.9 box mAP) using a YOLACT++ R-50 model.  Moreover, Mask R-CNN in fact has a slightly larger mAP difference (35.7 mask, 38.2 box), which suggests that the gap between the two methods lies in the relatively poor performance of our detector and not in our approach to generating masks. 

\paragraph{Quality of mask coefficients} 
As YOLACT produces masks by combining prototypes with mask coefficients, it would be nice to inspect the quality of those predicted coefficients. In order to do this, after training a YOLACT++ R-50 model, we freeze everything but the mask coefficient branch, and fine-tune only the coefficient predictor on the \emph{evaluation set}. With this setting, we only improve mask mAP from 33.7 to 33.9 even though we essentially have access to the ``optimal coefficients'' (i.e., fitting coefficients to test data given fixed prototypes). This shows that our predicted coefficients are very close to ``optimal coefficients'' and it is therefore a more promising direction to improve prototypes in order to minimize the gap between box and mask mAP.

%, which shows a good potential of our mask generating method.
%at the start of this section

%\daniel{It remains to be seen from future works how YOLACT fares with a more powerful backbone without regard to speed.}

%%%%%%%%%%%% CONCLUSION %%%%%%%%%%%%
% \vspace{-0.05in}
\section{Conclusion}
% \vspace{-0.05in}
We presented the first competitive single-stage real-time instance segmentation method.  The key idea is to predict mask prototypes and per-instance mask coefficients in parallel, and linearly combine them to form the final instance masks. Extensive experiments on MS COCO and Pascal VOC demonstrated the effectiveness of our approach and contribution of each component. We also analyzed the emergent behavior of our prototypes to explain how \methodname{}, even as an FCN, introduces translation variance for instance segmentation. Finally, with improvements to the backbone network, a better anchor design, and a fast mask re-scoring network, our \methodname{}++ showed a significant boost compared to the original framework while still running at real-time.

%which is usually considered to be translation invariant, can localize instances without explicitly adding modules which 

%\chong{We present the first single-stage real-time instance segmentation method with a competitive performance by predicting mask prototypes and per-instance mask coefficients in parallel followed by a simple linear combination to form the final instance masks. Extensive experiments have been done on MS COCO and Pascal VOC. We also conduct ablation studies to demonstrate the effectiveness and design reasoning of each component. In addition, we analyze the emergent behavior to explain how \methodname{} as a FCN, which is usually considered to be translation invariant, can localize instances without explicitly adding modules which introduce translation variance such as FCIS and Mask R-CNN. We also provide the box results and qualitative results in this paper. Finally, we upgrade the original \methodname{} with DCN, better anchor choice, and fast mask re-scoring network, which shows a significant boost and still runs at real-time.}

% use section* for acknowledgment
\ifCLASSOPTIONcompsoc
  % The Computer Society usually uses the plural form
  \section*{Acknowledgments}
\else
  % regular IEEE prefers the singular form
  \section*{Acknowledgment}
\fi

This work was supported in part by ARO YIP W911NF17-1-0410, NSF CAREER IIS-1751206, NSF IIS-1812850, AWS ML Research Award, Google Cloud Platform research credits, and XSEDE IRI180001. %and GPUs donated by NVIDIA. %The views and conclusions contained in this document are those of the authors and should not be interpreted as representing the official policies, either expressed or implied, of ARO or the U.S. Government. The U.S. Government is authorized to reproduce and distribute reprints for Government purposes notwithstanding any copyright notation herein.

% Can use something like this to put references on a page
% by themselves when using endfloat and the captionsoff option.
\ifCLASSOPTIONcaptionsoff
  \newpage
\fi

% references section
{
    \bibliographystyle{IEEEtran}
    \bibliography{IEEEabrv,YOLACT++}
}

% biography section
\newpage
\vspace{-30pt}
\begin{IEEEbiography}[{\includegraphics[width=1in,height=1.25in,clip,keepaspectratio]{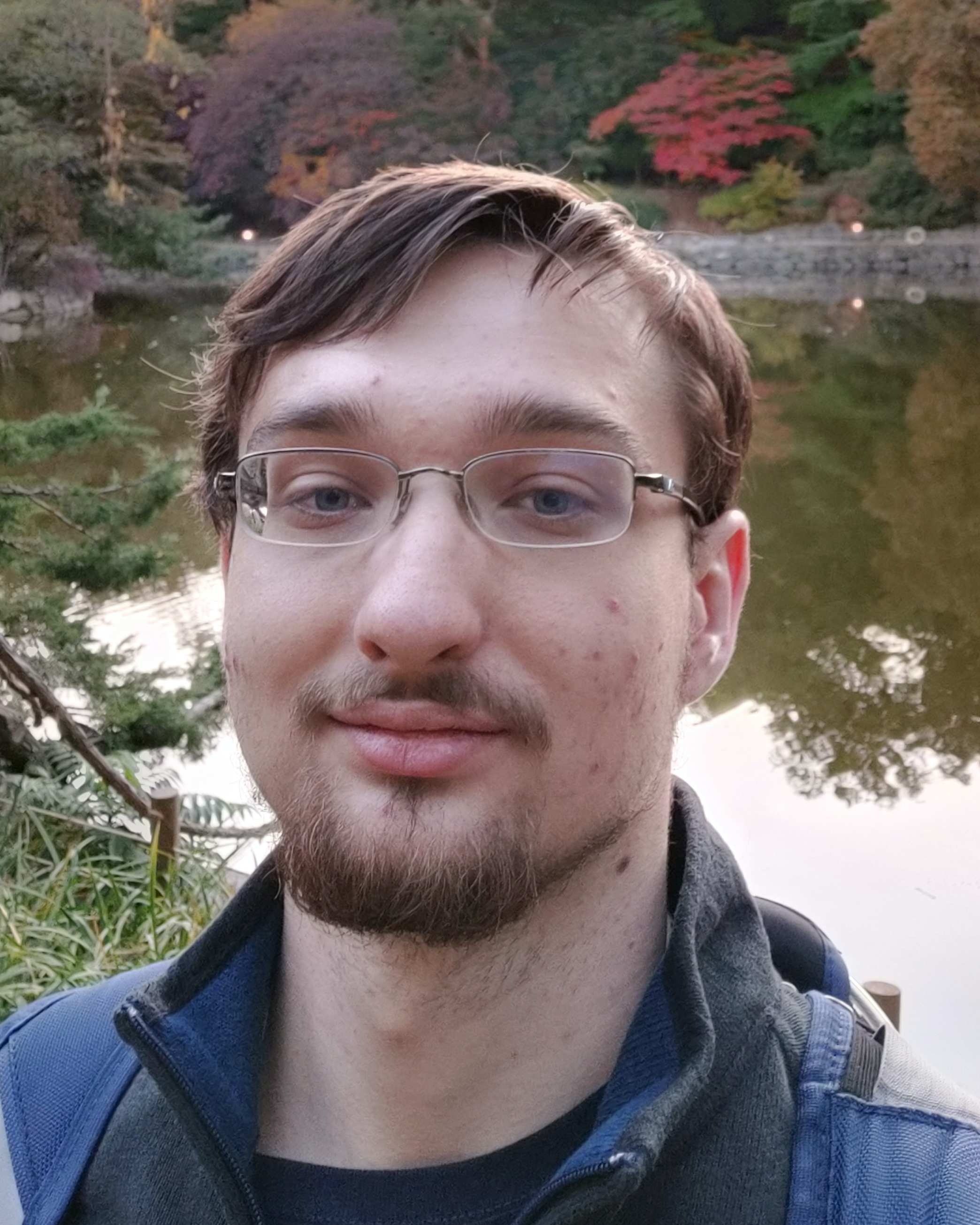}}]{Daniel Bolya}
graduated with a Bachelors of Science with Honors in 2019 from University of California, Davis. Daniel did most of his work for YOLACT while as an undergraduate and is currently pursuing a Ph.D. in Machine Learning at Georgia Institute of Technology. His interests lie broadly in the space of addressing core deficiencies in Machine Learning and Computer Vision, including speed, data usage, and generalizability.
\end{IEEEbiography}

% if you will not have a photo at all:
\vspace{-20pt}
\begin{IEEEbiography}[{\includegraphics[width=1in,height=1.25in,clip,keepaspectratio]{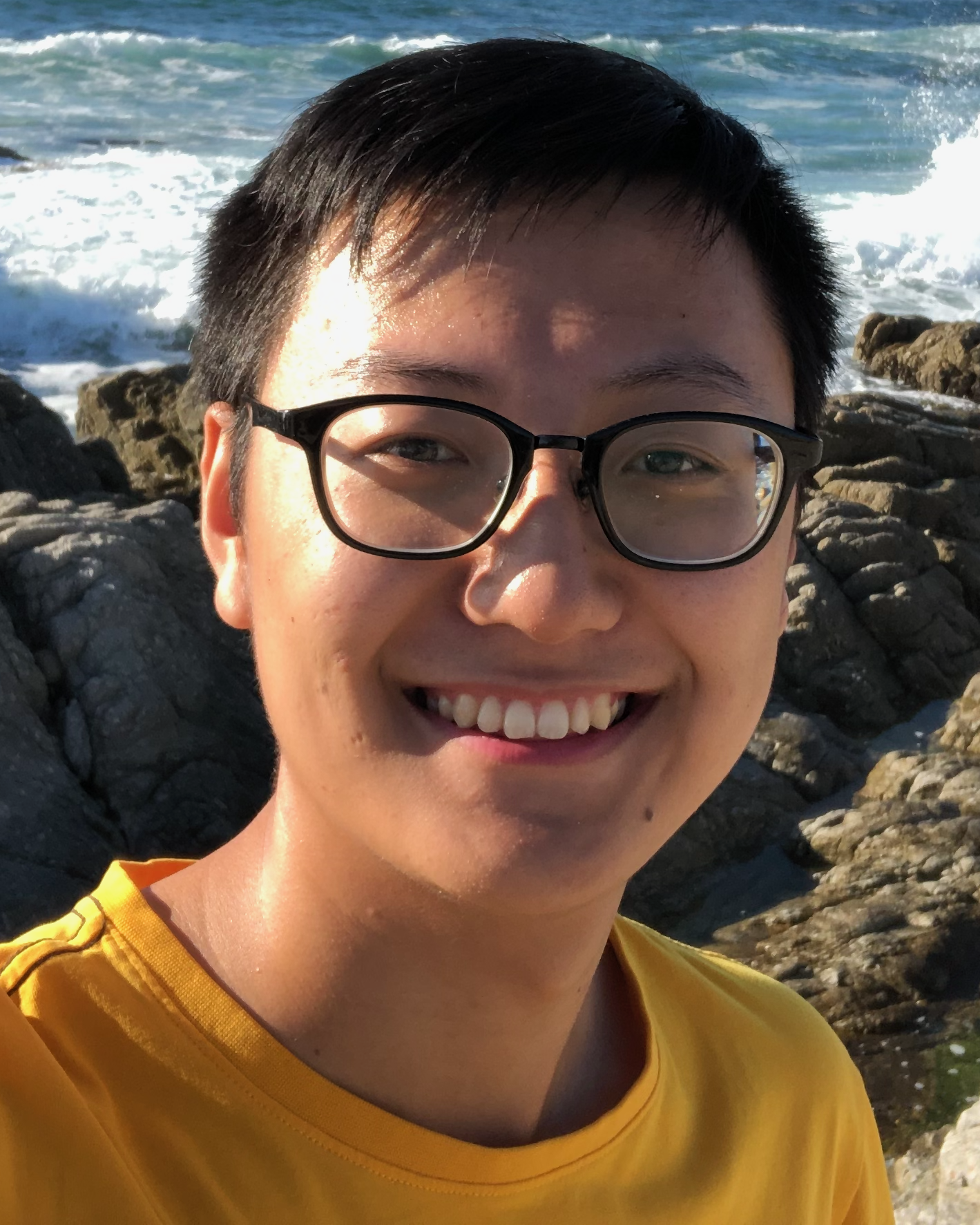}}]{Chong Zhou}
is a Master's student from the Department of Computer Science at the University of California, Davis. Prior to that, he completed his Bachelor's degree in Software Engineering from Nankai University, Tianjin, China. Chong is interested in Computer Vision and related problems in Machine Learning.
\end{IEEEbiography}

% insert where needed to balance the two columns on the last page with
% biographies
%\newpage
\vspace{-20pt}
\begin{IEEEbiography}[{\includegraphics[width=1in,height=1.25in,clip,keepaspectratio]{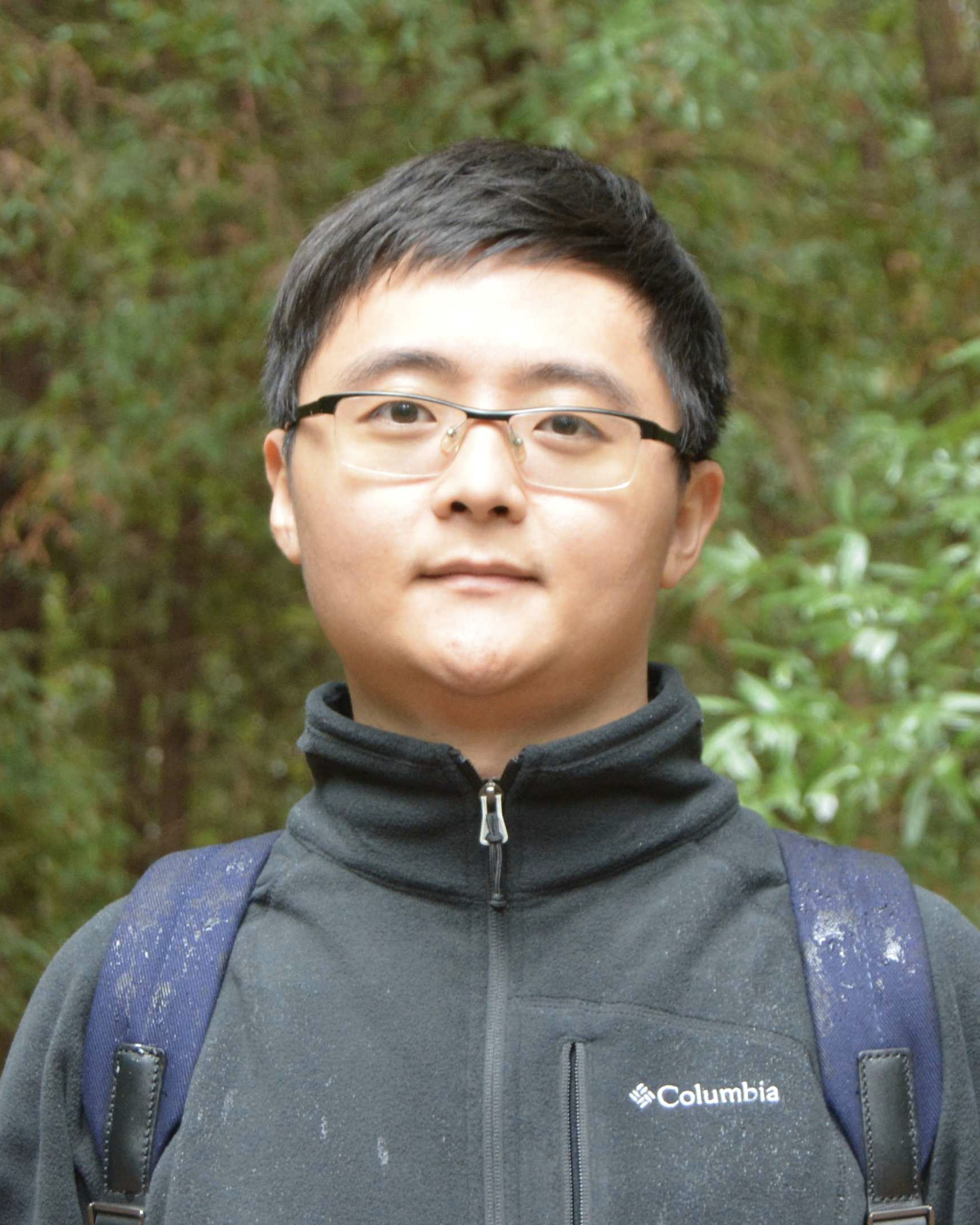}}]{Fanyi Xiao}
is a PhD candidate working in the Computer Vision Lab at the University of California, Davis. Before this, he obtained his Master's degree in Robotics from Carnegie Mellon University (Pittsburgh, USA) and his Bachelor's degree in Computer Science from Central South University (Changsha, China). He is broadly interested in deep learning for computer vision, and video understanding in particular. 
\end{IEEEbiography}

\vspace{-20pt}
\begin{IEEEbiography}[{\includegraphics[width=1in,height=1.25in,clip,keepaspectratio]{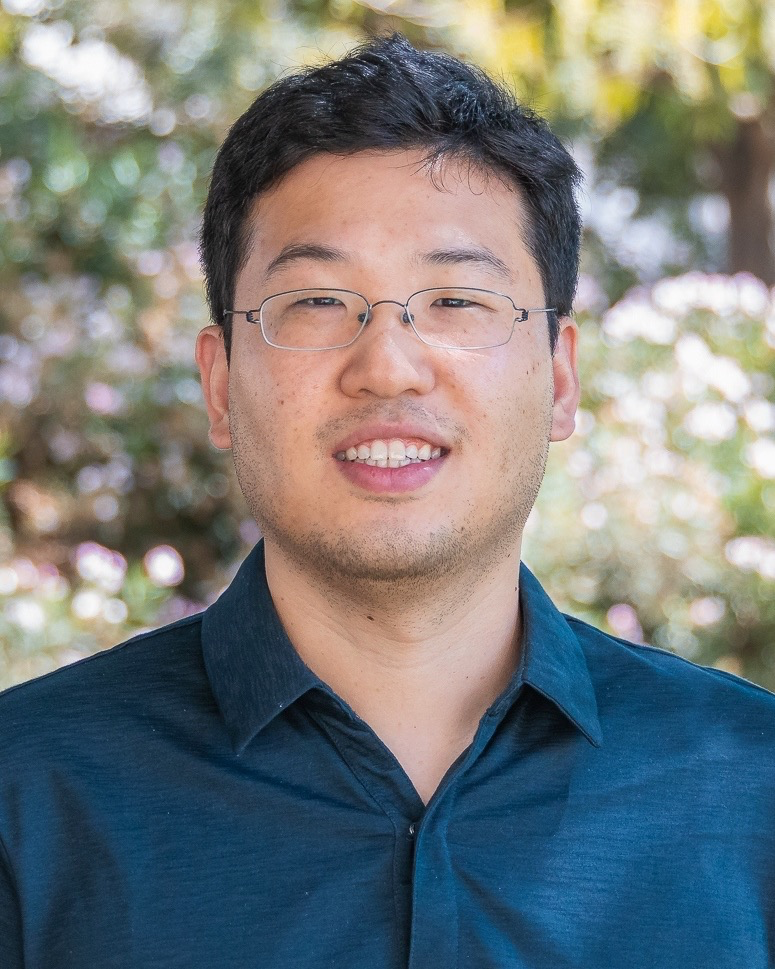}}]{Yong Jae Lee} is an Assistant Professor in the Department of Computer Science at the University of California, Davis. He received his Ph.D. from the University of Texas at Austin in 2012, and was a post-doc at Carnegie Mellon University (2012-2013) and UC Berkeley (2013-2014). He received his B.S. in Electrical Engineering from the University of Illinois at Urbana-Champaign in 2006. He is a recipient of the Army Research Office (ARO) Young Investigator Program (YIP) award, National Science Foundation (NSF) CAREER award, and UC Davis College of Engineering Outstanding Junior Faculty award.  His main research interests are in computer vision and machine learning.
\end{IEEEbiography}

\end{document}